\documentclass{article}

\usepackage{iclr2023_conference,times}

\definecolor{mydarkblue}{rgb}{0,0.08,0.45}
\usepackage[colorlinks=true,
    linkcolor=mydarkblue,
    citecolor=mydarkblue,
    filecolor=mydarkblue,
    urlcolor=mydarkblue]{hyperref}

\usepackage[utf8]{inputenc} 
\usepackage[T1]{fontenc}    
\usepackage{hyperref}       
\usepackage{url}            
\usepackage{booktabs}       
\usepackage{amsfonts}       
\usepackage{nicefrac}       
\usepackage{microtype}      
\usepackage{xcolor}         

\usepackage{pifont}
\newcommand{\ymark}{\ding{51}}%
\newcommand{\xmark}{\ding{55}}%
\usepackage{tikz}
\usepackage{amsmath}
\usepackage{amssymb}
\usepackage{wrapfig}
\usepackage{multirow}
\usepackage{tabularx}
\usepackage{adjustbox}
\usepackage{color, colortbl}
\usepackage{graphicx}
\usepackage{float} 
\usepackage{caption}
\usepackage{subcaption}

\usepackage{makecell}
\usepackage{caption}
\usepackage{enumerate}
\usepackage{bm}

\usepackage{tikz}
\usepackage{siunitx}
\usepackage{scalerel}

\usepackage{algorithm}
\usepackage{algorithmic}
\usepackage{soul}
\definecolor{myblue}{HTML}{0070C0}

\usepackage[marginal]{footmisc}

\newcommand{\x}{\mathbf{x}}

\newcommand{\y}{\mathbf{y}}
\newcommand{\z}{\mathbf{z}}

\newcommand{\dkl}{\mathbb{D}_{\rm{KL}}}

\usepackage{algorithm}
\usepackage{algorithmic}

\newcommand{\tikzsymbol}[1][circle, size=0.5cm]{\tikz[baseline=-0.5ex]\node[shape=#1,draw]{};}%

\definecolor{myred}{HTML}{C80728}

\definecolor{mygreen}{HTML}{C8ED28}

\definecolor{mygrad}{HTML}{C89628}

\definecolor{mColor2}{rgb}{0.95,0.95,0.95}

\title{
Energy-Based Test Sample Adaptation for \\Domain Generalization
}

\author{
Zehao Xiao\textsuperscript{1},
Xiantong Zhen\textsuperscript{1,2}\thanks{Currently with United Imaging Healthcare, Co., Ltd., China.}~,
Shengcai Liao\textsuperscript{2},
Cees G. M. Snoek\textsuperscript{1} \\
\textsuperscript{1}AIM Lab, University of Amsterdam \textsuperscript{2}Inception Institute of Artificial Intelligence}

\iclrfinalcopy
\begin{document}

\maketitle

\begin{abstract}
In this paper, we propose energy-based sample adaptation at test time for domain generalization. Where previous works adapt their models to target domains, we adapt the unseen target samples to source-trained models. To this end, we design a discriminative energy-based model, which is trained on source domains to jointly model the conditional distribution for classification and data distribution for sample adaptation. The model is optimized to simultaneously learn a classifier and an energy function. To adapt target samples to source distributions, we iteratively update the samples by energy minimization with stochastic gradient Langevin dynamics. Moreover, to preserve the categorical information in the sample during adaptation, we introduce a categorical latent variable into the energy-based model. The latent variable is learned from the original sample before adaptation by variational inference and fixed as a condition to guide the sample update. Experiments on six benchmarks for classification of images and microblog threads demonstrate the effectiveness of our proposal. 
\end{abstract}

\section{Introduction}

Deep neural networks are vulnerable to domain shifts and suffer from lack of generalization on test samples that do not resemble the ones in the training distribution \citep{recht2019imagenet,zhou2021domain,krueger2021out,shen2022association}.
To deal with the domain shifts, domain generalization has been proposed \citep{muandet2013domain,gulrajani2020search, cha2021swad}.
Domain generalization strives to learn a model exclusively on source domains in order to generalize well on unseen target domains.
The major challenge stems from the large domain shifts and the unavailability of any target domain data during training. 

To address the problem, domain invariant learning has been widely studied, e.g., \citep{motiian2017unified,zhao2020domain,nguyen2021domain}, based on the assumption that invariant representations obtained on source domains are also valid for unseen target domains. However, since the target data is inaccessible during training, it is likely an ``adaptivity gap'' \citep{dubey2021adaptive} exists between representations from the source and target domains. Therefore, recent works try to adapt the classification model with target samples at \textit{test time} by further fine-tuning model parameters~\citep{sun2020test,wang2021tent} or by introducing an extra network module for adaptation~\citep{dubey2021adaptive}. Rather than adapting the model to target domains, \cite{xiao2022learning} adapt the classifier for each sample at test time. Nevertheless, a single sample would not be able to adjust the whole model due to the large number of model parameters and the limited information contained in the sample. This makes it challenging for their method to handle large domain gaps. Instead, we propose to adapt each target sample to the source distributions, which does not require any fine-tuning or parameter updates of the source model.

In this paper, we propose energy-based test sample adaptation for domain generalization. The method is motivated by the fact that energy-based models \citep{hinton2002training,lecun2006tutorial} flexibly model complex data distributions and allow for efficient sampling from the modeled distribution by Langevin dynamics \citep{du2019implicit,welling2011bayesian}. Specifically, we define a new discriminative energy-based model as the composition of a classifier and a neural-network-based energy function in the data space, which are trained simultaneously on the source domains. The trained model iteratively updates the representation of each target sample by gradient descent of energy minimization through Langevin dynamics, which eventually adapts the sample to the source data distribution. The adapted target samples are then predicted by the classifier that is simultaneously trained in the discriminative energy-based model. For both efficient energy minimization and classification, we deploy the energy functions on the input feature space rather than the raw images.

Since Langevin dynamics tends to draw samples randomly from the distribution modeled by the energy function, it cannot guarantee category equivalence. To maintain the category information of the target samples during adaptation and promote better classification performance, we further introduce a categorical latent variable in our energy-based model. Our model learns the latent variable to explicitly carry categorical information by variational inference in the classification model. We utilize the latent variable as conditional categorical attributes like in compositional generation \citep{du2020compositional,nie2021controllable} to guide the sample adaptation to preserve the categorical information of the original sample. At inference time, we simply ensemble the predictions obtained by adapting the unseen target sample to each source domain as the final domain generalization result.

We conduct experiments on six benchmarks for classification of images and microblog threads to demonstrate the promise and effectiveness of our method for domain generalization\footnote{Code available:~\url{https://github.com/zzzx1224/EBTSA-ICLR2023}.}. 

\section{Methodology}

In domain generalization, we are provided source and target domains as non-overlapping distributions on the joint space $\mathcal{X} \times \mathcal{Y}$, where $\mathcal{X}$ and $\mathcal{Y}$ denote the input and label space, respectively. 
Given a dataset with $S$ source domains $\mathcal{D}_{s} {=} \left \{ D_{s}^{i} \right \}_{i=1}^{S}$ and $T$ target domains $\mathcal{D}_{t} {=} \left \{ D_{t}^{i}\right \}_{i=1}^{T}$, a model is trained only on $\mathcal{D}_{s}$ and required to generalize well on $\mathcal{D}_{t}$. 
Following the multi-source domain generalization setting \citep{li2017deeper, zhou2021domain}, we assume there are multiple source domains with the same label space to mimic good domain shifts during training.

In this work, we propose energy-based test sample adaptation, which adapts target samples to source distributions to tackle the domain gap between target and source data.
The rationale behind our model is that adapting the target samples to the source data distributions is able to improve the prediction of the target data with source models by reducing the domain shifts, as shown in Figure~\ref{fig1_intuition} (left). 
Since the target data is never seen during training, we mimic domain shifts during the training stage to learn the sample adaptation procedure. By doing so, the model acquires the ability to adapt each target sample to the source distribution at inference time.
In this section, we first provide a preliminary on energy-based models and then present our energy-based test sample adaptation.

\subsection{Energy-based model preliminary}
\label{background}

Energy-based models \citep{lecun2006tutorial} represent any probability distribution $p(\x)$ for $\x \in \mathbb{R}^D$ as $p_{\theta}(\x) = \frac{ {\rm exp} (-E_{\theta}(\x))}{Z_{\theta}}$, where $E_{\theta}(\x): \mathbb{R}^D \rightarrow \mathbb{R}$ is known as the energy function that maps each input sample to a scalar and $Z_{\theta} = \int {\rm exp} (-E_{\theta}(\x)) d \x$ denotes the partition function. 
However, $Z_{\theta}$ is usually intractable since it computes the integration over the entire input space of $\x$. Thus, we cannot train the parameter $\theta$ of the energy-based model by directly maximizing the log-likelihood $ {\rm log}~p_{\theta}(\x) = - E_{\theta}(x) - {\rm log} Z_{\theta}$.
Nevertheless, the log-likelihood has the derivative \citep{du2019implicit,song2021train} :
\begin{equation}
\label{contras_div}
    \frac{\partial {\rm log}~p_{\theta}(\x)}{\partial \theta} = \mathbb{E}_{p_d(\x)} \Big [ - \frac{\partial E_\theta(\x)}{\partial \theta} \Big ] + \mathbb{E}_{p_\theta(\x)} \Big [ \frac{\partial E_\theta(\x)}{\partial \theta} \Big ],
\end{equation}
where the first expectation term is taken over the data distribution $p_d(\x)$ and the second one is over the model distribution $p_\theta(\x)$.

The objective function in eq.~(\ref{contras_div}) encourages the model to assign low energy to the sample from the real data distribution while assigning high energy to those from the model distribution.
To do so, we need to draw samples from $p_\theta(\x)$, which is challenging and usually approximated by MCMC methods \citep{hinton2002training}. An effective MCMC method used in recent works \citep{du2019implicit, nijkamp2019learning, xiao2020vaebm, grathwohl2019your} is Stochastic Gradient Langevin Dynamics \citep{welling2011bayesian}, which simulates samples by
\begin{equation}
\label{langevin}
    \x^{i+1} = \x^{i} - \frac{\lambda}{2} \frac{\partial E_{\theta}(\x^{i})}{\partial \x^{i}} + \epsilon, ~~~~ s.t.,~~~~ \epsilon \sim \mathcal{N}(0, \lambda) 
\end{equation}
where $\lambda$ denotes the step-size and $\x^0$ is drawn from the initial distribution $p_0(\x)$, which is usually a uniform distribution \citep{du2019implicit,grathwohl2019your}.

Actually, maximizing  $ {\rm log}~p_{\theta}(\x)$ is equivalent to minimizing the KL divergence $\dkl (p_d(\x)||p_\theta(\x))$ \citep{song2021train}, which is alternatively achieved in \citep{hinton2002training} by minimizing contrastive divergence: 
\begin{equation}
\label{hintoncd}
    \dkl (p_d(\x)||p_\theta(\x)) - \dkl (q_\theta(\x)||p_\theta(\x)),
\end{equation}
where $q_\theta(\x) = \prod_\theta^t p_d(\x)$, representing $t$ sequential MCMC transitions starting from $p(\x)$ \citep{du2020improved} and minimizing eq.~(\ref{hintoncd}) is achieved by minimizing: 
\begin{equation}
\label{hintoncd2}
    \mathbb{E}_{p_d(\x)} [E_{\theta}(\x)] - \mathbb{E}_{\rm{stop\_grad}(q_\theta (\x))} [E_{\theta}(\x)] + \mathbb{E}_{q_\theta (\x)} [E_{\rm{stop\_grad}(\theta)}(\x)] + \mathbb{E}_{q_\theta (\x)} [{\rm log} q_\theta (\x)].
\end{equation}
Eq.~(\ref{hintoncd2}) avoids drawing samples from the model distribution $p_\theta(\x)$, which often requires an exponentially long time for MCMC sampling \citep{du2020improved}. 
Intuitively, $q_{\theta}(\x)$ is closer to $p_{\theta}(\x)$ than $p_d(\x)$, which guarantees that $\dkl (p_d(\x)||p_\theta(\x)) \geq \dkl (q_\theta(\x)||p_\theta(\x))$ and eq.~(\ref{hintoncd}) can only be zero when $p_{\theta}(\x)=p_d(\x)$.

\subsection{Energy-based test sample adaptation}
\label{sec3.1}

\begin{figure*}[t] 
\centering 
\centerline{\includegraphics[width=0.99\textwidth]{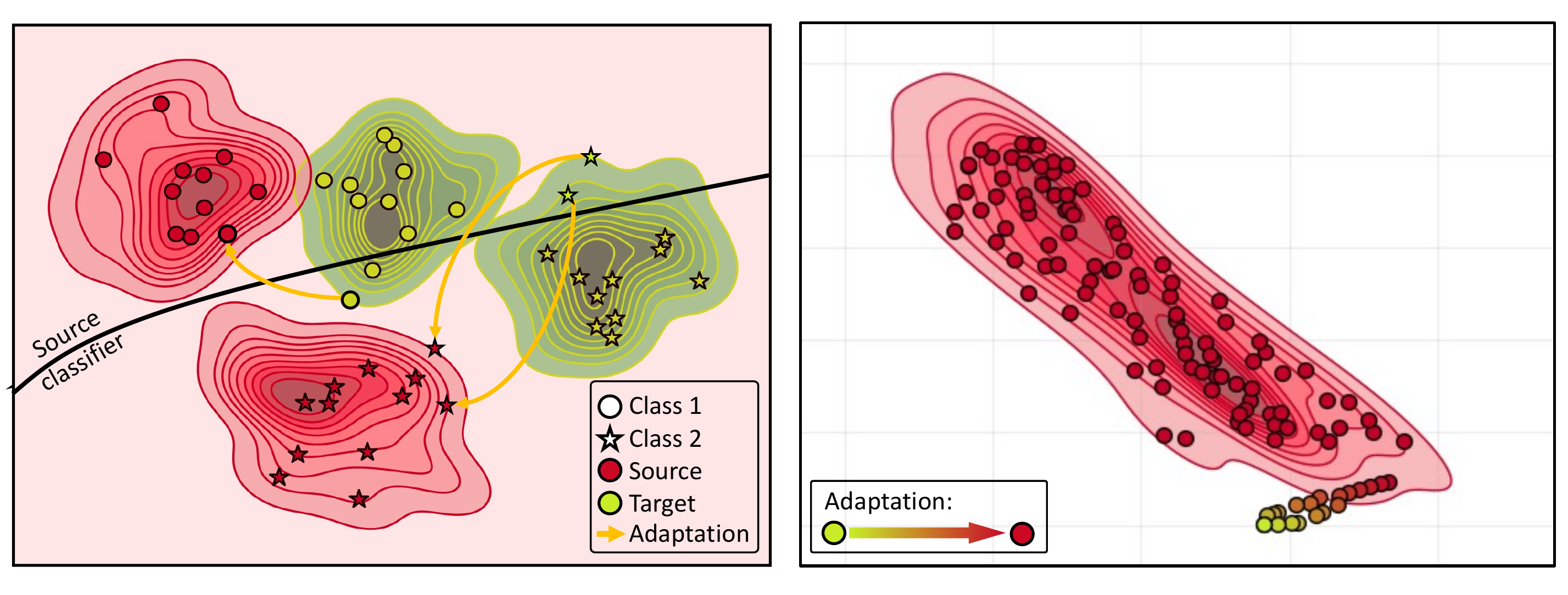}} 
\vspace{-2mm}
\caption{\textbf{Illustration of the proposed energy-based model.} It aims to adapt the target samples to the source distributions, which can be more accurately classified by the source domain classifier (left). With Langevin dynamics, each target sample is adapted iteratively to the source data distributions, which are represented as the gradient colors from green to red (right). Best viewed in color.
} 
\label{fig1_intuition}
\vspace{-4mm}
\end{figure*} 

We propose energy-based test sample adaption to tackle the domain gap between source and target data distributions. 
This is inspired by the fact that Langevin dynamics simulates samples of the distribution expressed by the energy-based model through gradient-based updates, with no restriction on the sample initialization if the sampling steps are sufficient \citep{welling2011bayesian,du2019implicit}.
We leverage this property to conduct test sample adaptation with Langevin dynamics by setting the target sample as the initialization and updating it iteratively. 
With the energy-based model of the source data distribution, as shown in Figure~\ref{fig1_intuition} (right), target samples are gradually updated towards the source domain and with sufficient update steps, the target sample will eventually be adapted to the source distribution.

\textbf{Discriminative energy-based model.}
We propose the discriminative energy-based model $p_{\theta,\phi}(\x,\y)$ on the source domain, which is constructed by a classification model and an energy function in the data space.
Note that $\x$ denotes the feature representations of the input image $I$, where
$\x$ is generated by a neural network backbone $\x{=}f_{\psi}(I)$.
Different from the regular energy-based models that generate data samples 
from uniform noise, our goal is to promote the discriminative task, i.e., the conditional distribution $p(\y|\x)$, which is preferred to be jointly modeled with the feature distributions $p(\x)$ of the input data.
Thus, the proposed energy-based model is defined on the joint space of $\mathcal{X} \times \mathcal{Y}$ to consider both the classification task and the energy function.
Formally, the discriminative energy-based model of a source domain is formulated as:
\begin{equation}
\label{basic}
    p_{\theta, \phi}(\x, \y) = p_{\phi}(\y|\x) \frac{ {\rm exp} (-E_{\theta}(\x))}{Z_{\theta, \phi}},
\end{equation}
where $p_{\phi}(\y|\x)$ denotes the classification model and $E_{\theta}(\x)$ is an energy function, which is nowadays implemented with neural networks.
Eq.~(\ref{basic}) enables the energy-based model to jointly model the feature distribution of input data and the conditional distribution on the source domains.
An unseen target sample $\x_t$ is iteratively adapted to the distribution of the source domain $D_s$ by Langevin dynamics update with the energy function $E_{\theta}(\x)$ and predicted by the classification model $p_{\phi}(\y|\x)$.

The model parameters $\theta$ and $\phi$ can be jointly optimized following eq.~(\ref{hintoncd}) by minimizing:
\begin{equation}
\dkl (p_d(\x, \y)||p_{\theta,\phi}(\x, \y)) - \dkl (q_{\theta,\phi} (\x, \y)||p_{\theta,\phi} (\x, \y)), 
\label{cd}
\end{equation}
which is derived as:
\begin{equation}
\begin{aligned}
\label{eq7:lall}
\mathcal{L} = 
& - \mathbb{E}_{p_{d}(\x, \y)} \big[ {\rm log}~p_{\phi}(\y|\x) \big] + \mathbb{E}_{p_{d}(\x, \y)} \big[ E_{\theta}(\x) \big] - \mathbb{E}_{\rm{stop\_grad}(q_{\theta, \phi}(\x, \y))} \big[E_{\theta}(\x) \big] \\ 
& + \mathbb{E}_{q_{\theta, \phi}(\x, \y)} \big[ E_{\rm{stop\_grad}(\theta)}(\x) - {{\rm log} ~ p_{\rm{stop\_grad}(\phi)}(\y|\x)} \big], 
\end{aligned}
\end{equation}
where $p_d(\x,\y)$ denotes the real data distribution and $q_{\theta, \phi}(\x, \y)=\prod_{\theta}^{t} p(\x, \y)$ denotes $t$ sequential MCMC samplings from the distribution expressed by the energy-based model similar to eq.~(\ref{hintoncd2}) \citep{du2020improved}.
We provide the detailed derivation in Appendix \ref{app:derive}.

In eq.~(\ref{eq7:lall}), the first term encourages to learn a discriminative classifier on the source domain. 
The second and third terms train the energy function to model the data distribution of the source domain by assigning low energy on the real samples and high energy on the samples from the model distribution. 
Different from the first three terms that directly supervise the model parameters $\theta$ and $\phi$, the last term stops the gradients of the energy function $E_{\theta}$ and classifier $\phi$ while back-propagating the gradients to the adapted samples $q_{\theta,\phi}(\x,\y)$.
Because of the stop-gradient, this term does not optimize the energy or log-likelihood of a given sample, but rather increases the probability of such samples with low energy and high log-likelihood under the modeled distribution.

Essentially, the last term trains the model $\theta$ to provide a variation for each sample that encourages its adapted version to be both discriminative on the source domain classifier and low energy on the energy function.
Intuitively, it supervises the model to learn the ability to preserve categorical information during adaptation and find a faster way to minimize the energy.

\textbf{Label-preserving adaptation with categorical latent variable.} 
Since the ultimate goal is to correctly classify target domain samples, it is necessary to maintain the categorical information in the target sample during the iterative adaptation process.
Eq.~(\ref{eq7:lall}) contains a supervision term that encourages the adapted target samples to be discriminative for the source classification models. 
However, as the energy function $E_{\theta}$ operates only in the $\mathcal{X}$ space and the sampling process of Langevin dynamics tends to result in random samples from the sampled distribution that are independent of the starting point, there is no categorical information considered during the adaptation procedure. 

To achieve label-preserving adaptation, we introduce a categorical latent variable $\z$ into the energy function to guide the adaptation of target samples to preserve the category information.
With the latent variable, the energy function $E_{\theta}$ is defined in the joint space of $\mathcal{X} \times \mathcal{Z}$. The categorical information contained in $\z$ will be explicitly incorporated into the iterative adaptation.
To do so, we define the energy-based model with the categorical latent variable as:
\begin{equation}
\begin{aligned}
\label{explicit}
    p_{\theta, \phi}(\x, \y) = \int p_{\theta, \phi}(\x, \y, \z) d\z
    = \int p_{\phi}(\y|\z, \x) p_{\phi}(\z|\x) \frac{ {\rm exp} (-E_{\theta}(\x|\z))}{Z_{\theta, \phi}}d\z,
\end{aligned}
\end{equation}
where $\phi$ denotes the parameters of the classification model that predicts $\z$ and $\y$ and $E_\theta$ denotes the energy function that models the distribution of $\x$ considering the information of latent variable $\z$.
$\z$ is trained to contain sufficient categorical information of $\x$ and serves as the conditional attributes that guide the adapted samples $\x$ preserving the categorical information. 
Once obtained from the original input feature representations $\x$, $\z$ is fixed and taken as the input of the energy function together with the updated $\x$ in each iteration. 
Intuitively, when $\x$ is updated from the target domain to the source domain via Langevin dynamics, $\z$ helps it preserve the classification information contained in the original $\x$, without introducing additional information.

To learn the latent variable $\z$ with more categorical information,
we estimate $\z$ by variational inference and design a variational posterior $q_\phi(\z|\mathbf{d}_{\x})$, where 
$\mathbf{d}_{\x}$ is the average representation of samples from the same category as $\x$ on the source domain. 
Therefore, $q_\phi(\z|\mathbf{d}_{\x})$ can be treated as a probabilistic prototypical representation of a class. 
By incorporating $q_\phi(\z|\mathbf{d}_{\x})$ into eq.~(\ref{explicit}), we obtain the lower bound of the log-likelihood:
\begin{equation}
\begin{aligned}
\label{explicit1}
    {\rm log}~p_{\theta, \phi}(\x, \y) 
    & \geq \mathbb{E}_{q_{\phi}} [{\rm log}~p_{\phi}(\y| \z,\x) - E_{\theta}(\x, \z) - {\rm log}~Z_{\theta,\phi}]  + \dkl [q_{\phi}(\z|\mathbf{d}_{\x}) || p_{\phi}(\z|\x)].
\end{aligned}
\end{equation}
Note that in eq.~(\ref{explicit1}), the categorical latent variable $\z$ is incorporated into both the classification model $p_{\phi}(\y| \z,\x)$ and the energy function $E_{\theta}(\x|\z)$. 
The energy function contains both data information and categorical information. 
During the Langevin dynamics update for sample adaptation, the latent variable provides categorical information in each iteration, which enables the adapted target samples to be discriminative.

By incorporating eq.~(\ref{explicit1}) into eq.~(\ref{cd}), we derive the objective function with the categorical latent variable as:
\begin{equation}
\begin{aligned}
\label{eq10explicitall}
    \mathcal{L}_{f} 
    & = \mathbb{E}_{p_d(\x, \y)} \Big [\mathbb{E}_{q_{\phi}(\z)} [- {\rm log}~p_{\phi}(\y| \z,\x)] + \dkl [q_{\phi}(\z|\mathbf{d_{\x}}) || p_{\phi}(\z|\x)]\Big ] + \mathbb{E}_{q_{\phi} (\z)} \Big[ \mathbb{E}_{p_d(\x)} [ E_{\theta}(\x|\z)] \\ 
     & - \mathbb{E}_{\rm{stop\_grad}(q_{\theta}(\x))} [E_{\theta}(\x, \z)] \Big ] 
     + \mathbb{E}_{q_{\theta}(\x)} \Big[ \mathbb{E}_{q_{\rm{stop\_grad}(\phi)}(\z)} \big[  E_{\rm{stop\_grad}(\theta)} (\x, \z) \\
     & - {\rm log}~ p_{\rm{stop\_grad}(\phi)} (\y|\z, \x) \big ] -\dkl [q_{\rm{stop\_grad} (\phi)} (\z|\mathbf{d}_{\x}) || p_{\rm{stop\_grad}(\phi)}(\z|\x)] \Big],
\end{aligned}
\end{equation}
where $p_d(\x)$ and $q_{\theta}(\x)$ denote the data distribution and the $t$ sequential MCMC samplings from the energy-based distribution of the source domain $D_s$.
Similar to eq.~(\ref{eq7:lall}), the first term trains the classification model on the source data. The second term trains the energy function to model the source data distribution. The last term is conducted on the adapted samples to supervise the adaptation procedure.
The complete derivation is provided in Appendix \ref{app:derive}.
An illustration of our model is shown in Figure~\ref{illustration}.
We also provide the complete algorithm in Appendix \ref{app:algo}.

\begin{figure*}[t] 
\centering 
\centerline{\includegraphics[width=0.99\textwidth]{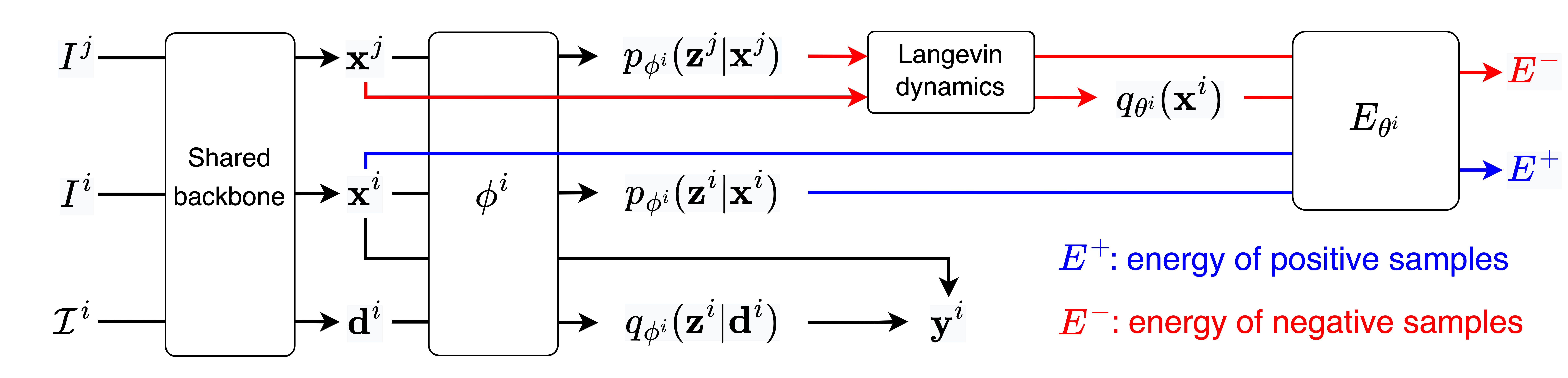}}
\caption{\textbf{Overall process of the proposed sample adaptation by discriminative energy-based model.} In each iteration, we train the classification model $\phi^{i}$ and energy function $E_{\theta^{i}}$ of one  source domain $D^{i}$. 
$I^i$ and $I^j$ denote the images from domains $D^{i}$ and $D^{j}$, respectively.
$\mathcal{I}^{i}$ denotes one batch of images, which generate the center features $\mathbf{d}^{i}$. The energy function $E_{\theta^{i}}$ is trained by using $\mathbf{x}^{i}$ as positive samples and adapted samples $q_{\mathbb{\theta}^{i}}(\mathbf{x}^{i})$ generated by $\mathbf{x}^{j}$ from other domains as negative samples. The adaptation is achieved by Langevin dynamics of $E_{\theta^{i}}$.
During inference, the target samples are adapted by Langevin dynamics of the energy function $E_{\theta}$ of each source domain and then predicted by eq.~(\ref{ensemble}).} 
\label{illustration}
\vspace{-4mm}
\end{figure*} 

\textbf{Ensemble inference.}
Since the target data is inaccessible during training, we train the specific parameters $\theta$ and $\phi$ to model each source distribution by adapting the samples from other source domains to the current source distribution.
In each iteration, we train the energy-based model $\theta^i$  of one randomly selected source domain $D^i_s$. The adapted samples generated by samples $\x^j, \x^k$ from the other source domains $D^j_s, D^k_s$ are used as the negative samples while $\x^i$ as the positive samples to train the energy-based model.
During inference, the target sample is adapted to each source distribution with the specific energy function and predicted by the specific classifier.
After that, we combine the predictions of all source domain models to obtain the final prediction:
\begin{equation}
\label{ensemble}
    p(\y_t) = \frac{1}{S} \sum^{S}_{i=1} \frac{1}{N} \sum^{N}_{n=1} p_{\phi^i}(\y|\z^n, \x)~~~~~~~~ \z^n \sim p(\z^n|\x_t), \x \sim p_{\theta^i}(\x).
\end{equation}
Here $\phi^i$ and $\theta^i$ denote the domain specific classification model and energy function of domain $D^i_s$.
Note that since the labels of the $\x^t$ are unknown, $\mathbf{d}_{\x^t}$ in eq.~(\ref{eq10explicitall}) is not available during inference. Therefore, we draw $\z^n$ from the prior distribution $p(\z^n|\x_t)$, where $\x_t$ is the original target sample without any adaptation.
With fixed $\z^n$, $\x \sim p_{\theta^i}(\x)$ are drawn by Langevin dynamics as in eq.~(\ref{langevin}) with the target samples $\x^t$ as the initialization sample.
$p_{\theta^i}(\x)$ denotes the distributions modeled by the energy function $E_{\theta^i}(\x|\z^n)$.
Moreover, to be efficient, {the feature extractor $\psi$ for obtaining feature representations $\x$} is shared by all source domains and only the energy functions and classifiers are domain specific.
We deploy the energy-based model on feature representations for lighter neural networks of the domain-specific energy functions and classification models. 

\section{Experiments}

\textbf{Datasets.}
We conduct our experiments on five widely used datasets for domain generalization, \textbf{PACS} \citep{li2017deeper}, \textbf{Office-Home} \citep{venkateswara2017deep}, \textbf{DomainNet} \citep{peng2019moment}, and \textbf{Rotated MNIST and Fashion-MNIST}.
Since we conduct the energy-based distribution on the feature space, our method can also handle other data formats. Therefore, we also evaluate the method on \textbf{PHEME} \citep{zubiaga2016analysing}, a dataset for natural language processing.

PACS consists of 9,991 images of seven classes from four domains, i.e., \textit{photo}, \textit{art-painting}, \textit{cartoon}, and \textit{sketch}. We use the same training and validation split as \citep{li2017deeper} and follow their ``leave-one-out'' protocol. 
Office-Home also contains four domains, i.e., \textit{art}, \textit{clipart}, \textit{product}, and \textit{real-world}, which totally have 15,500 images of 65 categories. 
DomainNet is more challenging since it has six domains i.e., \textit{clipart}, \textit{infograph}, \textit{painting}, \textit{quickdraw}, \textit{real}, \textit{sketch}, with 586,575 examples of 345 classes.
We use the same experimental protocol as PACS. 
We utilize the Rotated MNIST and Fashion-MNIST datasets by following the settings in Piratla et al. \citep{piratla2020efficient}. 
The images are rotated from $0^\circ$ to $90^\circ$ in intervals of $15^\circ$, covering seven domains. 
We use the domains with rotation angles from $15^\circ$ to $75^\circ$ as the source domains, and images rotated by $0^\circ$ and $90^\circ$ as the target domains. 
PHEME is a dataset for rumour detection. There are a total of 5,802 tweets labeled as rumourous or non-rumourous from 5 different events, i.e., \textit{Charlie Hebdo}, \textit{Ferguson}, \textit{German  Wings}, \textit{Ottawa Shooting}, and \textit{Sydney Siege}. Same as PACS, we evaluate our methods on PHEME also by the ``leave-one-out'' protocol. 

\textbf{Implementation details.}
We evaluate on PACS and Office-Home with both a ResNet-18 and ResNet-50 \citep{he2016deep} and on DomainNet with a ResNet-50. The backbones are pretrained on ImageNet \citep{deng2009imagenet}.
On PHEME we conduct the experiments based on a pretrained DistilBERT \citep{sanh2019distilbert}, following \citep{wright2020transformer}. 
{To increase the number of sampling steps and sample diversity of the energy functions, we introduce a replay buffer $\mathcal{B}$ that stores the past updated samples from the modeled distribution \citep{du2019implicit}.}
The details of the models and hyperparameters are provided in Appendix \ref{app:imp}.

\begin{table}[t]
\centering
\caption{\textbf{Benefit of energy-based test sample adaptation.} Experiments on PACS using a ResNet-18 averaged over five runs. Optimized by eq.~(\ref{eq7:lall}), our model improves after adaptation. With the latent variable (eq.~\ref{eq10explicitall}) performance improves further, both before and after adaptation.}
\centering
\label{ablations}
	\resizebox{0.99\columnwidth}{!}{%
		\setlength\tabcolsep{4pt} 
\begin{tabular}{lcccccc}
\toprule 
& Adaptation & \textbf{Photo}    & \textbf{Art-painting }        & \textbf{Cartoon}        & \textbf{Sketch}         & \textit{Mean}       \\ \midrule
\multirow{2}*{Without latent variable (eq.~\ref{eq7:lall})} & \xmark & 94.73 \scriptsize{$\pm$0.22} & 78.66 \scriptsize{$\pm$0.59} & 78.24 \scriptsize{$\pm$0.71} & 78.34 \scriptsize{$\pm$0.62} & 82.49 \scriptsize{$\pm$0.26} \\
~ & \ymark  & 94.59 \scriptsize{$\pm$0.16} & 80.45 \scriptsize{$\pm$0.52} & 79.98 \scriptsize{$\pm$0.51} & \textbf{79.23} \scriptsize{$\pm$0.32} & 83.51 \scriptsize{$\pm$0.30} \\
\midrule
\multirow{2}*{With latent variable (eq.~\ref{eq10explicitall})} & \xmark  & 95.12 \scriptsize{$\pm$0.41} & 79.79 \scriptsize{$\pm$0.64} & 79.15 \scriptsize{$\pm$0.37} & \textbf{79.28} \scriptsize{$\pm$0.82} & 83.33 \scriptsize{$\pm$0.43} \\
~ & \ymark  & \textbf{96.05} \scriptsize{$\pm$0.37} & \textbf{82.28} \scriptsize{$\pm$0.31} & \textbf{81.55} \scriptsize{$\pm$0.65} & \textbf{79.81} \scriptsize{$\pm$0.41} & \textbf{84.92} \scriptsize{$\pm$0.59} \\
\bottomrule
\end{tabular}
}
\end{table}

\textbf{Benefit of energy-based test sample adaptation.}
We first investigate the effectiveness of our energy-based test sample adaptation in Table~\ref{ablations}. Before adaptation, we evaluate the target samples directly by the classification model of each source domain and ensemble the predictions. After the adaptation to the source distributions, the performance of the target samples improves, especially on the \textit{art-painting} and \textit{cartoon} domains, demonstrating the benefit of the iterative adaptation by our energy-based model. The results of models with latent variable, i.e., trained by eq.~(\ref{eq10explicitall}), are shown in the last two rows. 
With the latent variable, the performance is improved both before and after adaptation, which shows the benefit of incorporating the latent variable into the classification model.
The performance improvement after adaptation is also more prominent than without the latent variable, demonstrating the effectiveness of incorporating the latent variable into the energy function. 

\begin{figure*}[t] 
\centering 
\centerline{\includegraphics[width=0.99\textwidth]{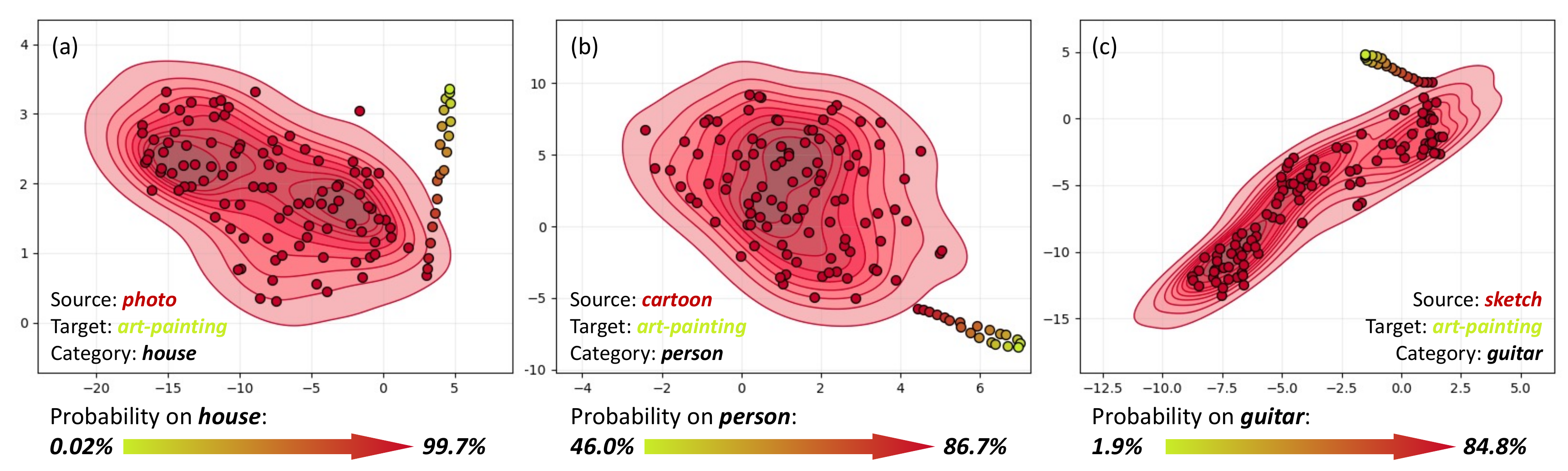}} 
\caption{\textbf{Iterative adaptation of target samples.} We adapt the samples from the target domain (\textit{art-painting}) to different source domains (\textit{photo}, \textit{cartoon}, and \textit{sketch} in (a), (b), (c)).
Each subfigure shows the adaptation of one target sample. The adaptation procedure of the target samples is represented by the gradient color from green to red. 
In each figure, the target sample has the same label as the source data, but is mispredicted due to domain shifts.
During adaptation, the target samples gradually approach the source distributions and eventually get correct predictions.} 
\label{visualization2a}
\vspace{-3mm}
\end{figure*} 

\textbf{Effectiveness of iterative test sample adaptation by Langevin dynamics.}
We visualize the iterative adaptation of the target samples in Figure~\ref{visualization2a}. 
In each subfigure, the target and source samples have the same label.
The visualization shows that the target samples gradually approach the source data distributions during the iterative adaptation by Langevin dynamics.
After adaptation, the predictions of the target samples on the source domain classifier also become more accurate.
For instance, in Figure~\ref{visualization2a} (a), the target sample of the \textit{house} category is predicted incorrectly, with a probability of \textit{house} being only 0.02\%.
After adaptation, the probability becomes 99.7\%, which is predicted correctly. More visualizations, including several failure cases, are provided in Appendix \ref{app:vis}.

\begin{wrapfigure}{r}{0.5\linewidth}
\centering
\includegraphics[width=0.95\linewidth]{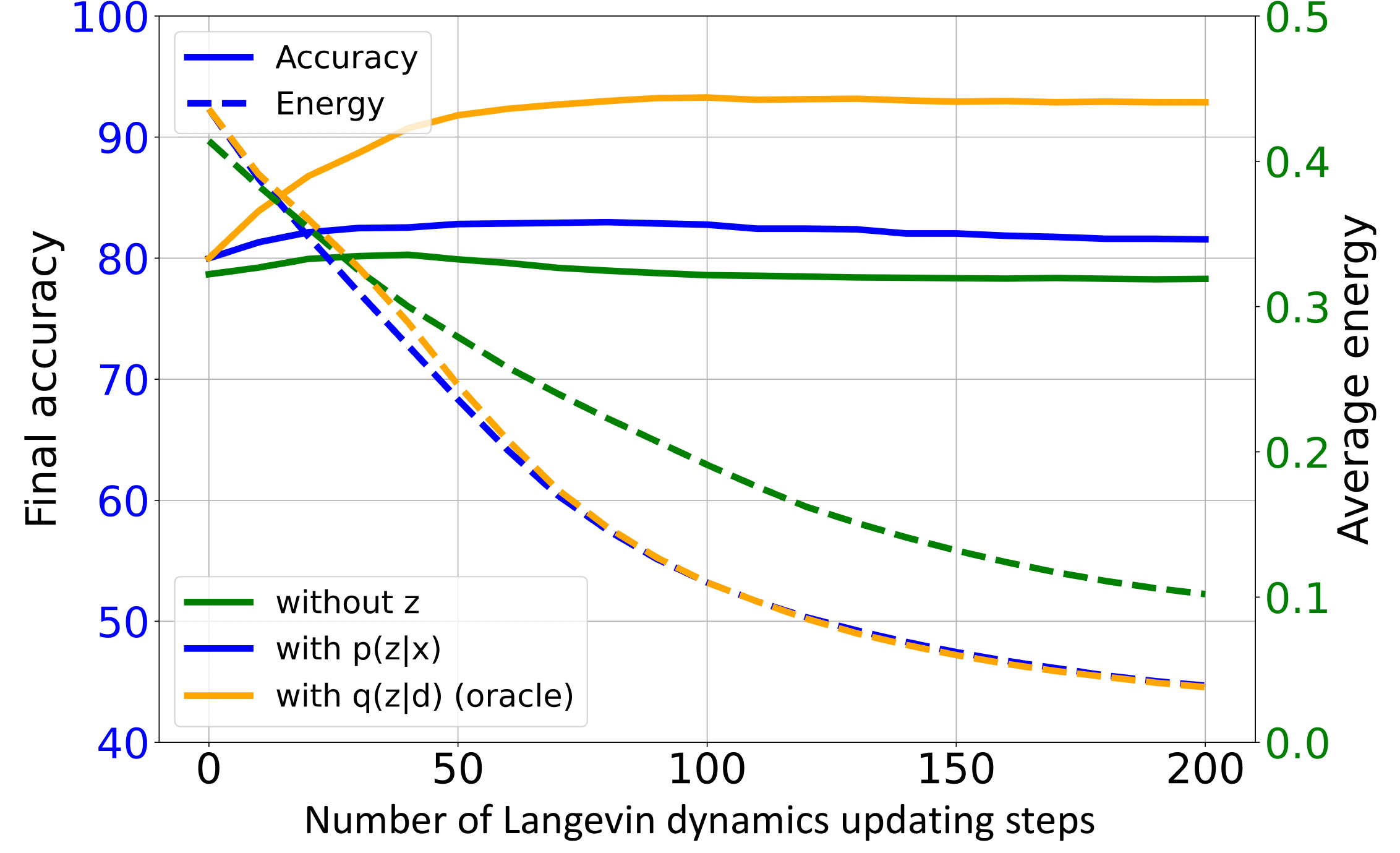}
\caption{
\textbf{Adaptation with different Langevin dynamics steps.} As the number of steps increases, energy decreases while accuracy increases. When the number of steps is too large, the accuracy without $\z$ or with $p(\z|\x)$ drops slightly while the accuracy with $q(\z|\mathbf{d}_{\x})$ is more stable and better.
}
\label{accene}
\vspace{-2mm}
\end{wrapfigure}

\textbf{Adaptation with different Langevin dynamics steps.}
We also investigate the effect of the Langevin dynamics step numbers during adaptation.
Figure~\ref{accene} shows the variety of the average energy and accuracy of the target samples adapted to the source distributions with different updating steps.
The experiments are conducted on PACS with ResNet-18. The target domain is \textit{art-painting}.
With the step numbers less than 80, the average energy decreases consistently while the accuracy increases along with the increased number of updating steps, showing that the target samples are getting closer to the source distributions. 
When the step numbers are too large, the accuracy will decrease as the number of steps increases. We attribute this to $\z_t$ having imperfect categorical information, since it is approximated during inference from a single target sample $\x_t$ only. In this case, the label information would not be well preserved in $\x_t$ during the Langevin dynamics update, which causes an accuracy drop with a large number of updates. 
To demonstrate this, we conduct the experiment by replacing $p(\z_t)$ with $q_{\phi}(\z|\mathbf{d}_\x)$ during inference. $\mathbf{d}_\x$ is the class center of the same class as $\x_t$. Therefore, $q_{\phi}(\z|\mathbf{d}_\x)$ contains categorical information that is closer to the ground truth label. We regard this as the oracle model. 
As expected, the oracle model performs better as the number of steps increases and reaches stability after 100 steps. We also show the results without $\z$. We can see the performance and stability are both worse, which again demonstrates that $\z$ helps preserve label information in the target samples during adaptation. 
{Moreover, the energy without conditioning on $\z$ is higher. The reason can be that without conditioning on $\z$, there is no guidance of categorical information during sample adaptation. In this case, the sample can be adapted randomly by the energy-based model, regardless of the categorical information. This can lead to the conflict to adapt the target features to different categories of the source data, slowing down the decline of the energy.} 
We provide more analyses of $\z_t$ in Appendix \ref{app:results}.
In addition, the training and test time cost is also larger as the step number increases, the comparisons and analyses are also provided in Appendix \ref{app:results}. 

\begin{table}[t]
\begin{center}
\vspace{-2mm}
\caption{\textbf{Comparisons on image and text datasets.} Our method achieves the best mean accuracy for all datasets, independent of the backbone. {Larger adaptation steps (i.e., 50) lead to better performance.}
}
\label{pacsoff}
	\resizebox{0.99\columnwidth}{!}{%
		\setlength\tabcolsep{4pt} 
\begin{tabular}{lllllll}
\toprule
~ & \multicolumn{2}{c}{\textbf{PACS}} & \multicolumn{2}{c}{\textbf{Office-Home}} & \textbf{DomainNet} & \textbf{PHEME} \\
\cmidrule(lr){2-3} \cmidrule(lr){4-5} \cmidrule(lr){6-6} \cmidrule(lr){7-7}
 & ResNet-18 & ResNet-50 & ResNet-18 & ResNet-50 & ResNet-50 & DistilBERT\\ \midrule
\cite{iwasawa2021test} & 81.40 & 85.10  & 57.00 & 68.30 & - & - \\
\cite{zhou2020learning}   & 82.83 & 84.90 & 65.63 & 67.66 & - & -\\
\cite{gulrajani2020search}  & - & 85.50 & - & 66.50 & 40.90 & - \\
\cite{wang2021tent}  & 83.09 & 86.23 & 64.13  & 67.99 & - & 75.8 \scriptsize{$\pm$0.23}\\
\cite{dubey2021adaptive}  & - &  & - & 68.90 & 43.90 & - \\
\cite{xiao2022learning} & 84.15 & 87.51 & 66.02  & 71.07 & - & 76.1 \scriptsize{$\pm$0.21}\\  
\midrule
\textbf{\textit{This paper w/o adaptation}}  & 83.33 \scriptsize{$\pm$0.43} & 86.05 \scriptsize{$\pm$0.37} & 65.01 \scriptsize{$\pm$0.47} & 70.44 \scriptsize{$\pm$0.25} & 42.90 \scriptsize{$\pm$0.34} & 75.4 \scriptsize{$\pm$0.13}\\ 

\textbf{\textit{This paper w/ adaptation (10 steps)}}  
 & 84.25 \scriptsize{$\pm$0.48} & 87.05 \scriptsize{$\pm$0.26} & 65.73 \scriptsize{$\pm$0.32} & 71.13 \scriptsize{$\pm$0.43} & 43.75 \scriptsize{$\pm$0.49} & 76.0 \scriptsize{$\pm$0.16}\\ 

\textbf{\textit{This paper w/ adaptation (20 steps)}}  
   & \textbf{84.92} \scriptsize{$\pm$0.59} & 87.70 \scriptsize{$\pm$0.28} & 66.31 \scriptsize{$\pm$0.21} & \textbf{72.07} \scriptsize{$\pm$0.38} & \textbf{44.66} \scriptsize{$\pm$0.51} & 76.5 \scriptsize{$\pm$0.18}\\ 

\textbf{\textit{This paper w/ adaptation (50 steps)}}  
  & \textbf{85.10} \scriptsize{$\pm$0.33} & \textbf{88.12} \scriptsize{$\pm$0.25} & \textbf{66.75} \scriptsize{$\pm$0.21} & \textbf{72.25} \scriptsize{$\pm$0.32} & \textbf{44.98} \scriptsize{$\pm$0.43} & \textbf{76.9} \scriptsize{$\pm$0.16}\\ 
\bottomrule
\end{tabular}
}
\end{center}
\vspace{-4mm}
\end{table}

\textbf{Comparisons.}
PACS, Office-Home, and DomainNet are three widely used benchmarks in domain generalization.
We conduct experiments on PACS and Office-Home based on both ResNet-18 and ResNet-50 and experiments on DomainNet based on ResNet-50.
As shown in Table~\ref{pacsoff}, our method achieves competitive and even the best overall performance in most cases. 
Moreover, our method performs better than most of the recent test-time adaptation methods \citep{iwasawa2021test,wang2021tent,dubey2021adaptive}, which fine-tunes the model at test time with batches of target samples.
{By contrast, we strictly follow the setting of domain generalization. We only use the source data to train the classification and energy-based models during training. At test time, we do our sample adaptation and make predictions on each individual target sample by just the source-trained models. 
Our method is more data efficient at test time, avoiding the problem of data collection per target domain in real-world applications. Despite the data efficiency during inference, our method is still comparable and sometimes better, especially on datasets with more categories, e.g., Office-Home and DomainNet.}
Compared with the recent work by \cite{xiao2022learning}, our method is at least competitive and often better.
To show the generality of our method, we also conduct experiments on the natural language processing dataset PHEME. The dataset is a binary classification task for rumour detection. The results in Table~\ref{pacsoff} show similar conclusions as the image datasets.

Table~\ref{pacsoff} also demonstrates the effectiveness of our sample adaptation. For each dataset and backbone, the proposed method achieves a good improvement after adaptation by the proposed discriminative energy-based model.
{For fairness, the results without adaptation are also obtained by ensemble predictions of the source-domain-specific classifiers.
Moreover, larger steps (i.e., 50) lead to better performance. The improvements of adaptation with 50 steps are slight. Considering the trade-off of computational efficiency and performance, we set the step number as 20 in our paper.}
We provide detailed comparisons, results on rotated MNIST and Fashion-MNIST datasets, as well as more experiments on the latent variable, corruption datasets, and analyses of the ensemble inference method in Appendix \ref{app:results}.

\section{Related work}


\textbf{Domain generalization.}
One of the predominant methods is domain invariant learning \citep{muandet2013domain,ghifary2016scatter,motiian2017unified,seo2020learning,zhao2020domain,xiao2021bit,mahajan2021domain,nguyen2021domain,phung2021learning,shi2021gradient}.
\cite{muandet2013domain} and \cite{ghifary2016scatter} learn domain invariant representations by matching the moments of features across source domains. 
Li et al. \citep{li2018domainb} further improved the model by learning conditional-invariant features.
Recently, \cite{mahajan2021domain} introduced causal matching to model within-class variations for generalization.
\cite{shi2021gradient} provided a gradient matching to encourage consistent gradient directions across domains.
\cite{arjovsky2019invariant} and \cite{ahuja2021invariance} proposed on invariant risk minimization to learn an invariant classifier.
Another widely used methodology is domain augmentation \citep{shankar2018generalizing,volpi2018generalizing,qiao2020learning, zhou2020learning,zhou2021mix, yao2022improving}, which generates more source domain data to simulate domain shifts during training. 
\cite{zhou2021mix} proposed a data augmentation on the feature space by mixing the feature statistics of instances from different domains. 
Meta-learning-based methods have also been studied for domain generalization \citep{li2018metalearning,balaji2018metareg,dou2019domain,du2020metanorm,bui2021exploiting,du2021hierarchical}. 
\cite{li2018metalearning} introduced the model agnostic meta-learning \citep{finn2017model} into domain generalization. 
\cite{du2020learning} proposed the meta-variational information bottleneck for domain-invariant learning. 

\textbf{Test-time adaptation and source-free adaptation.}
Recently, adaptive methods have been proposed to better match the source-trained model and the target data at test time
\citep{sun2020test,li2020model,d2019learning,pandey2021generalization,iwasawa2021test,dubey2021adaptive,zhang2021adaptive}. 
Test-time adaptation \citep{sun2020test, wang2021tent, liu2021ttt++, zhou2021training} fine-tunes (part of) a network trained on source domains by batches of target samples.
\cite{xiao2022learning} proposed single-sample generalization that adapts a model to each target sample under a meta-learning framework. 
There are also some source-free domain adaptation methods \citep{liang2020we,yang2021exploiting,dong2021confident,liang2021source} that adapt the source-trained model on only the target data. 
These methods follow the domain adaptation settings to fine-tune the source-trained model by the entire target set. 
By contrast, we do sample adaptation at test time but strictly follow the domain generalization settings. In our method, no target sample is available during the training of the models. At test time, each target sample is adapted to the source domains and predicted by the source-trained model individually, without fine-tuning the models or requiring large amounts of target data.

\textbf{Energy-based model.} The energy-based model is a classical learning framework \citep{ackley1985learning,hinton2002training,hinton2006unsupervised,lecun2006tutorial}. 
Recently, \citep{xie2016theory,nijkamp2019learning,nijkamp2020anatomy,du2019implicit,du2020improved, xie2022tale} further extend the energy-based model to high-dimensional data using contrastive divergence and Stochastic Gradient Langevin dynamics.
Different from most of these works that only model the data distributions, some recent works model the joint distributions \citep{grathwohl2019your, xiao2020vaebm}.
In our work, we define the joint distribution of data and label to promote the classification of unseen target samples in domain generalization, and further incorporate a latent variable to incorporate the categorical information into the Langevin dynamics procedure.
Energy-based models for various tasks have been proposed, e.g., image generation \citep{du2020compositional,nie2021controllable}, 
out-of-distribution detection \citep{liu2020energy}, and anomaly detection \citep{dehaene2020iterative}. 
Some methods also utilize energy-based models for domain adaptation \citep{zou2021unsupervised,xie2021active,kurmi2021domain}. 
Different from these methods, we focus on domain generalization and utilize the energy-based model to express the source domain distributions without any target data during training. 

\section{Conclusion and Discussions}
In this paper, we propose a discriminative energy-based model to adapt the target samples to the source data distributions for domain generalization.
The energy-based model is designed on the joint space of input, output, and a latent variable, which is constructed by a domain specific classification model and an energy function. 
With the trained energy-based model, the target samples are adapted to the source distributions through Langevin dynamics and then predicted by the classification model.
Since we aim to prompt the classification of the target samples, the model is trained to achieve label-preserving adaptation by incorporating the categorical latent variable.
We evaluate the method on six image and text benchmarks. The results demonstrate its effectiveness and generality.
We have not tested our approach beyond image and text classification tasks, but since our sample adaptation is conducted on the feature space, it should be possible to extend the method to other complex tasks based on feature representations.

Compared with recent model adaptation methods, our method does not need to adjust the model parameters at test time, which requires batches of target samples to provide sufficient target information.
This is more data efficient and challenging at test time, therefore the training procedure is more involved with complex optimization objectives.
One limitation of our proposed method is the iterative adaptation requirement for each target sample, which introduces an extra time cost at both training and test time. The problem can be mitigated by speeding up the energy minimization with optimization techniques during Langevin dynamics, e.g.,  Nesterov momentum \citep{nesterov1983method}, or by exploring one-step methods for sample adaptation. We leave these explorations for future work.

\section*{Acknowledgment}
This work is financially supported by the Inception Institute of Artificial Intelligence, the University of Amsterdam and the allowance 
Top consortia for Knowledge and Innovation (TKIs) from the Netherlands Ministry of Economic Affairs and Climate Policy.

\bibliography{iclr2023_conference}
\bibliographystyle{iclr2023_conference}
\newpage

\appendix

\section{Derivations}
\label{app:derive}
\textbf{Derivation of energy-based sample adaptation.} 
Recall our discriminative energy-based model 
\begin{equation}
p_{\theta, \phi}(\x, \y) = p_{\phi}(\y|\x) \frac{ {\rm exp} (-E_{\theta}(\x))}{Z_{\theta, \phi}},   
\end{equation}
where $Z_{\theta, \phi} = \int p_{\phi}(\y|\x) {\rm exp} (-E_{\theta}(\x)) d\x d\y$ is the partition function.
$\phi$ and $\theta$ denote the parameters of the classifier and energy function, respectively. 
To jointly train the parameters, we minimize the contrastive divergence proposed by \cite{hinton2002training}:
\begin{equation}
\begin{aligned}
\label{lossall}
\mathcal{L} = \dkl[p_d(\x, \y)||p_{\theta, \phi}(\x, \y)] - \dkl[q_{\theta, \phi}(\x, \y)||p_{\theta, \phi}(\x, \y)],
\end{aligned}
\end{equation}
where $p_d(\x,\y)$ denotes the real data distribution and $q_{\theta, \phi}(\x, \y)=\prod_{\theta}^{t} p(\x, \y)$ denotes $t$ sequential MCMC samplings from the distribution expressed by the energy-based model \citep{du2020improved}.
The gradient of the first term with respect to $\theta$ and $\phi$ is
\begin{equation}
\begin{aligned}
\label{term1grad}
\mathbf{\nabla}_{\theta, \phi} \dkl[p_d(\x, \y)||p_{\theta, \phi}(\x, \y)]
 & =  \mathbf{\nabla}_{\theta, \phi} \mathbb{E}_{p_{d}(\x, \y)} \big[{\rm log}~\frac{p_{d}(\x, \y)}{p_{\theta, \phi}(\x, \y)} \big] \\
 & =  \mathbb{E}_{p_{d}(\x, \y)} \big[\mathbf{\nabla}_{\theta, \phi} {\rm log}~p_{d}(\x, \y) - \mathbf{\nabla}_{\theta, \phi}  {\rm log}~p_{\theta, \phi}(\x, \y) \big] \\
 & =  \mathbb{E}_{p_{d}(\x, \y)} \big[ - \mathbf{\nabla}_{\theta, \phi}  {\rm log}~p_{\theta, \phi}(\x, \y) \big], \\
\end{aligned}
\end{equation}
while the gradient of the second term is
\begin{equation}
\begin{aligned}
\label{term2grad}
& \mathbf{\nabla}_{\theta, \phi} \dkl[q_{\theta, \phi}(\x, \y)||p_{\theta, \phi}(\x, \y)] \\
 = & \mathbf{\nabla}_{\theta, \phi} \mathbb{E}_{q_{\theta, \phi}(\x,\y)} \big[{\rm log}~\frac{q_{\theta, \phi}(\x, \y)}{p_{\theta, \phi}(\x, \y)} \big] \\
= & \mathbf{\nabla}_{\theta, \phi} q_{\theta, \phi}(\x,\y) \mathbf{\nabla}_{q_{\theta, \phi}} \dkl[q_{\theta, \phi}(\x, \y)||p_{\theta, \phi}(\x, \y)] + \mathbb{E}_{q_{\theta, \phi}(\x, \y)} \big[ - \mathbf{\nabla}_{\theta, \phi}  {\rm log}~p_{\theta, \phi}(\x, \y) \big]. \\
\end{aligned}
\end{equation}
Combining eq.~(\ref{term1grad}) and eq.~(\ref{term2grad}), we have the overall gradient as:
\begin{equation}
\begin{aligned}
\label{allgrad}
\mathbf{\nabla}_{\theta, \phi} \mathcal{L}_{all}
= & - (\mathbb{E}_{p_{d}(\x, \y)} \big[\mathbf{\nabla}_{\theta, \phi}  {\rm log}~p_{\theta, \phi}(\x, \y) \big] - \mathbb{E}_{q_{\theta, \phi}(\x, \y)} \big[\mathbf{\nabla}_{\theta, \phi}  {\rm log}~p_{\theta, \phi}(\x, \y) \big]\\
& + \mathbf{\nabla}_{\theta, \phi} q_{\theta, \phi}(\x,\y) \mathbf{\nabla}_{q_{\theta, \phi}} \dkl[q_{\theta, \phi}(\x, \y)||p_{\theta, \phi}(\x, \y)]).
\end{aligned}
\end{equation}
For the first two terms, the gradient can be further derived to 
\begin{equation}
\begin{aligned}
\label{term1grad2}
& \mathbb{E}_{p_{d}(\x, \y)} \big[\mathbf{\nabla}_{\theta, \phi}  {\rm log}~p_{\theta, \phi}(\x, \y) \big] - \mathbb{E}_{q_{\theta, \phi}(\x, \y)} \big[\mathbf{\nabla}_{\theta, \phi}  {\rm log}~p_{\theta, \phi}(\x, \y) \big] \\
= & \mathbb{E}_{p_{d}(\x, \y)} \big[\mathbf{\nabla}_{\theta, \phi} ({\rm log}~p_{\phi}(\y|\x) - E_{\theta}(\x) - {\rm log}~Z_{\theta, \phi})\big]  \\
- & \mathbb{E}_{q_{\theta, \phi}(\x, \y)} \big[\mathbf{\nabla}_{\theta, \phi} ({\rm log}~p_{\phi}(\y|\x) - E_{\theta}(\x) - {\rm log}~Z_{\theta, \phi})\big].
\end{aligned}
\end{equation}
Moreover, $\mathbf{\nabla}_{\theta, \phi} {\rm log}~Z_{\theta, \phi}$ can be written as the expectation $\mathbb{E}_{p_{\theta, \phi}(\x, \y)} [\mathbf{\nabla}_{\phi} {\rm log}~p_{\phi}(\y|\x) -\mathbf{\nabla}_{\theta} E_{\theta}(\x)]$ \citep{song2021train,xiao2020vaebm}, which is therefore canceled out in eq.~(\ref{term1grad2}) \citep{hinton2002training}.
We then have the loss function for the first two terms as
\begin{equation}
\begin{aligned}
\label{fterm1}
\mathcal{L}_{1}
 = \mathbb{E}_{p_{d}(\x, \y)} \big[E_{\theta}(\x)  - {\rm log}~p_{\phi}(\y|\x) \big] - \mathbb{E}_{q_{\theta, \phi}(\x, \y)} \big[E_{\theta}(\x) - {\rm log}~p_{\phi}(\y|\x) \big].
\end{aligned}
\end{equation}
Furthermore, we have the loss function 
\begin{equation}
\begin{aligned}
\label{fterm2}
\mathcal{L}_{2}
& = \mathbb{E}_{q_{\theta, \phi}(\x, \y)} \big[{\rm log} \frac{q_{\theta, \phi}(\x, \y)}{p_{stop\_grad(\theta, \phi)}(\x, \y)} \big] \\
& =  - ~ \mathbb{E}_{q_{\theta, \phi}(\x, \y)} \big[{{\rm log} p_{stop\_grad(\phi)}(\y|\x)} - E_{stop\_grad(\theta)}(\x) - {\rm log} Z_{stop\_grad(\theta, \phi)}\big] \\
& ~~~~~ + \mathbb{E}_{q_{\theta, \phi}(\x, \y)} \big[{\rm log} q_{\theta, \phi}(\x, \y) \big],
\end{aligned}
\end{equation}
which has the same gradient as the last term in eq.~(\ref{allgrad}) \citep{du2020improved}.
The $stop\_grad$ here means that we do not backpropagate the gradients to update the parameters by the corresponding forward functions. Thus, these parameters can be treated as constants.

Since the gradient of $\theta$ and $\phi$ is stopped in ${\rm log}~Z_{stop\_grad(\theta, \phi)}$, we treat it as a constant independent of $q_{\theta, \phi}(\x, \y)$ and therefore remove it from the eq.~(\ref{fterm2}).
In addition, the term $\mathbb{E}_{q_{\theta, \phi}(\x, \y)} \big[{\rm log}~p_{\phi}(\y|\x) \big]$ in eq.~(\ref{fterm1}) encourages wrong prediction of the updated samples from $q_{\theta, \phi}(\x, \y)$, which goes against our goal of promoting classification by adapting target samples.
The term $\mathbb{E}_{q_{\theta, \phi}(\x, \y)} \big[{\rm log} q_{\theta, \phi}(\x, \y) \big]$ in eq.~(\ref{fterm2}) can be treated as a negative entropy of $q_{\theta, \phi}(\x, \y)$, which is always negative and hard to estimate.
Therefore, we remove these two terms in the final loss function by applying an upper bound of the combination of eq.~(\ref{fterm1}) and eq.~(\ref{fterm2}) as:
\begin{equation}
\begin{aligned}
\label{lall}
\mathcal{L} = 
& - \mathbb{E}_{p_{d}(\x, \y)} \big[ {\rm log}~p_{\phi}(\y|\x) \big] + \mathbb{E}_{p_{d}(\x, \y)} \big[ E_{\theta}(\x) \big] - \mathbb{E}_{\rm{stop\_grad}(q_{\theta, \phi}(\x, \y))} \big[E_{\theta}(\x) \big] \\ 
& + \mathbb{E}_{q_{\theta, \phi}(\x, \y)} \big[ E_{\rm{stop\_grad}(\theta)}(\x) - {{\rm log} ~ p_{\rm{stop\_grad}(\phi)}(\y|\x)} \big].
\end{aligned}
\end{equation}

\textbf{Energy-based sample adaptation with categorical latent variable.}
To keep the categorical information during sample adaptation, we introduce a categorical latent variable $\z$ into our discriminative energy-based model, which is defined as $p_{\theta, \phi}(\x, \y) = \int p_{\theta, \phi}(\x, \y, \z) d\z = \int p_{\phi}(\y|\z, \x) p_{\phi}(\z|\x) \frac{ {\rm exp} (-E_{\theta}(\x|\z))}{Z_{\theta, \phi}}d\z$.
We optimize the parameters $\theta$ and $\phi$ also by the contrastive divergence $\dkl[p_d(\x, \y)||p_{\theta, \phi}(\x, \y)] - \dkl[q_{\theta, \phi}(\x, \y)||p_{\theta, \phi}(\x, \y)]$, which has similar gradient as eq.~(\ref{term1grad}) and eq.~(\ref{term2grad}).
The latent variable $\z$ is estimated by variational inference, leading to a lower bound of ${\rm log}~ p_{\theta, \phi}(\x, \y) \geq \mathbb{E}_{q_{\phi}(\z)} [{\rm log}~p_{\phi}(\y|\z,\x) - E_{\theta}(\x|\z) - {\rm log}~Z_{\theta,\phi}]  + \dkl [q_{\phi}(\z|\mathbf{d}_{\x}) || p_{\phi}(\z|\x)]$.
We obtain the final loss function of the contrastive divergence in a similar way as eq.~(\ref{lall}) by estimating the gradient and remove the terms that are hard to estimate or conflict with our final goal.
The final objective function is:
\begin{equation}
\begin{aligned}
\label{explicitall}
\mathcal{L}_{f} 
    & = \mathbb{E}_{p_d(\x, \y)} \Big [\mathbb{E}_{q_{\phi}(\z)} [- {\rm log}~p_{\phi}(\y| \z,\x)] + \dkl [q_{\phi}(\z|\mathbf{d_{\x}}) || p_{\phi}(\z|\x)]\Big ] + \mathbb{E}_{q_{\phi} (\z)} \Big[ \mathbb{E}_{p_d(\x)} [ E_{\theta}(\x|\z)] \\ 
     & - \mathbb{E}_{\rm{stop\_grad}(q_{\theta}(\x))} [E_{\theta}(\x, \z)] \Big ]
     + \mathbb{E}_{q_{\theta}(\x)} \Big[ \mathbb{E}_{q_{\rm{stop\_grad}(\phi)}(\z)} \big[  E_{\rm{stop\_grad}(\theta)} (\x, \z) \\
     & - {\rm log}~ p_{\rm{stop\_grad}(\phi)} (\y|\z, \x) \big ] -\dkl [q_{\rm{stop\_grad} (\phi)} (\z|\mathbf{d}_{\x}) || p_{\rm{stop\_grad}(\phi)}(\z|\x)] \Big].
\end{aligned}
\end{equation}

\section{Algorithm}
\label{app:algo}
We provide the detailed training and test algorithm of our energy-based sample adaptation in Algorithm~\ref{algo:1}.

\begin{algorithm}[t]
\algsetup{linenosize=\tiny}
\small
\caption{Energy-based sample adaptation}
\label{algo:1}
\begin{algorithmic}
\STATE \underline{TRAINING TIME}
\STATE {\textbf{Require:}} Source domains $\mathcal{D}_{s} {=} \left \{ D_{s}^{i} \right \}_{i=1}^{S}$ each with joint distribution $p_{d_s^i}{(I, \y})$ of input image and label.
\STATE {\textbf{Require:}} Learning rate $\mu$; iteration numbers $M$; step numbers $K$ and step size $\lambda$ of the energy function.
\STATE Initialize pretrained backbone $\psi$; $\phi^i, \theta^i, \mathcal{B}^i=\varnothing$ for each source domain $D_{s}^{i}$. 
\FOR{\textit{iter} in $M$}
\FOR{$D_{s}^{i}$ in $\mathcal{D}_{s}$}
\STATE Sample datapoints $\{(I^i, \y^i)\} \sim p_{d_s^i}{(I, \y})$; $\{(I^j, \y^j)\} \sim \{p_{d_s^j}{(I, \y})\}_{j \neq i}$ or $\mathcal{B}$ with 50\% probability.
\STATE Feature representations {$\x^i=f_{\psi}(I^i), \x^j=f_{\psi}(I^j)$}
\FOR{$k$ in $K$}
\STATE $\x^j_k \leftarrow \x^j_{k-1} - \mathbf{\nabla}_{\x} E_{\theta^i}(\x^j_{k-1}|\z^j) + \omega$, ~~~ $\z^j \sim q_{\phi^i}(\z^j|\mathbf{d}_{\x^j}), ~~ \omega \sim \mathcal{N}(0, \sigma)$.
\ENDFOR
\STATE $q_{\theta^i}(\x) \leftarrow \x^j_k$, ~~~~~ $p_d(\x) \leftarrow p_{d_s^i}(\x^i)$.
\STATE 
$(\psi, \phi^i) \leftarrow (\psi, \phi^i) -\lambda \nabla_{\psi, \phi^i} \mathcal{L}_{f}(p_d(\x))$ 
~~~~$\theta^i \leftarrow \theta^i -\lambda \nabla_{\theta^i} \mathcal{L}_{f}(p_d(\x), q_{\theta^i}(\x))$. 
\STATE $\mathcal{B}^i \leftarrow \mathcal{B}^i \cup \x^j_k$
\ENDFOR
\ENDFOR
\STATE {\hrulefill}
\STATE \underline{TEST TIME}
\STATE {\textbf{Require}}: 
Target images $I_t$ from the target domain; trained backbone $\psi$; and domain-specific model $\phi^i, \theta^i$ for each source domain in $\left \{ D_{s}^{i} \right \}_{i=1}^{S}$.
\STATE Input feature representations {$\x_t=f_{\psi}(I_t)$}.
\FOR{$i$ in $\left \{1, \dots, S \right \}$}
\FOR{$k$ in $K$}
\STATE $\x_{t, k} \leftarrow \x_{t, k-1} - \mathbf{\nabla}_{\x} E_{\theta^i}(\x_{t, k-1}|\z_t) + \omega$, ~~~ $\z_t \sim p_{\phi^i}(\z_t|\x_t), ~~ \omega \sim \mathcal{N}(0, \sigma)$.
\ENDFOR
\STATE $\y^i_t=p_{\phi^i}(\y_t|\x_{t, k}, \z_t)$
\ENDFOR
\STATE \textbf{return} $\y_{t} = \frac{1}{S} \sum^{S}_{i=1} \y_t^i$.
\end{algorithmic}
\end{algorithm}

\section{Datasets and Implementation details}
\label{app:imp}

\textbf{Model.}
To be efficient, we train a shared backbone for all source domains while a domain-specific classifier and a neural-network-based energy function for each source domain.
{The feature extractor backbone is a basic Residual Network without the final fully connected layer (classifier).}
Both the prior distribution $p_{\phi}(\z|\x)$ and posterior distribution $q_{\phi} (\z|\mathbf{d}_{\x})$ of the latent variable $\z$ are generated by a neural network $\phi$ that consists of four fully connected layers with ReLU activation, which outputs the mean and variance of the distribution.
The last layer of $\phi$ outputs both the mean and standard derivation of the distribution $p_{\phi}(\z|\x)$ and $q_{\phi} (\z|\mathbf{d}_{\x})$ for further Monte Carlo sampling.
The dimension of $\z$ is the same as the feature representations $\x$, e.g., 512 for ResNet-18 and 2048 for ResNet-50.
$\mathbf{d}_{\x}$ is obtained by the center features of the batch of samples that have the same categories as the current sample $\x$ in each iteration.

Deployed on feature representations, the energy function consists of three fully connected layers with two dropout layers.
The latent variable $\z$ is incorporated into the energy function by concatenating with the feature representation $\x$.
The input dimension is doubled of the output feature of the backbone, i.e., 1024 for ResNet-18 and 4096 for ResNet-50.
We use the \textit{swish} function as activation in the energy functions \citep{du2020improved}. 
The final output of the EBM is a scalar, which is processed by a sigmoid function following \cite{du2020improved} to bound the energy to the region $[0, 1]$ and improve the stability during training.
During training, we introduce a replay buffer $\mathcal{B}$ to store the past updated samples from the modeled distribution \citep{du2019implicit}. By sampling from $\mathcal{B}$ with 50\% probability, we can initialize the negative samples with either the sample features from other source domains or the past Langevin dynamics procedure. This can increase the number of sampling steps and the sample diversity.

\begin{table}[t]
\caption{\textbf{Implementation details} of our method per dataset and backbone.
}
\centering
\label{param}
	\resizebox{0.8\columnwidth}{!}{%
		\setlength\tabcolsep{4pt} 
\begin{tabular}{lcccc}
\toprule
Dataset & Backbone & Backbone learning rate  & Step size  & Number of steps \\ \midrule
\multirow{2}*{PACS} & ResNet-18  & 0.00005 & 50 & 20 \\ 
~ & ResNet-50 & 0.00001  & 50 & 20 \\ \midrule
\multirow{2}*{Office-Home} & ResNet-18   & 0.00001 & 100 & 20 \\
~ & ResNet-50   & 0.00001 & 100 & 20 \\ \midrule
Rotated MNIST & ResNet-18  & 0.00005 & 50 & 20 \\ \midrule
Fashion-MNIST & ResNet-18  & 0.00005 & 50 & 20 \\  \midrule
PHEME & DistilBERT  & 0.00003 & 40 & 20 \\
\bottomrule
\end{tabular}
}
\vspace{-4mm}
\end{table}

\textbf{Training details and hyperparameters.}
We evaluate on PACS with both a ResNet-18 and ResNet-50 pretrained on ImageNet. 
We use Adam optimization and train for 10,000 iterations with a batch size of 128. We set the learning rate to 0.00005 for ResNet-18, 0.00001 for ResNet-50, and 0.0001 for the energy-based model and classification model. We use 20 steps of Langevin dynamics sampling to adapt the target samples to source distributions, with a step size of 50. 
We set the number of Monte Carlo sampling $N$ in eq.~(\ref{ensemble}) as 10 for PACS.
Most of the experimental settings on Office-Home are the same as on PACS. The learning rate of the backbone is set to 0.00001 for both ResNet-18 and ResNet-50.
The number of Monte Carlo sampling is 5.
For fair comparison, we evaluate the rotated MNIST and Fashion-MNIST with ResNet-18, following \citep{piratla2020efficient}. The other settings are also the same as PACS.
On PHEME we conduct the experiments based on a pretrained DistilBERT. We set the learning rate as 0.00003 and use 20 steps of Langevin dynamics with a step size of 20.

We train all models on an NVIDIA Tesla V100 GPU for 10,000 iterations.
The learning rates of the backbone are different for different datasets as shown in Table~\ref{param}.
The learning rates of the domain-specific classifiers and energy functions are both set to 0.0001 for all datasets.
For each source domain, we randomly select 128 samples as a batch to train the backbone and classification model.
We also select 128 samples from the other source domains together with the current domain samples to train the domain-specific energy function.
We use a replay buffer with 500 feature representations and 
apply spectral normalization
on all weights of the energy function \citep{du2019implicit}.
We use random noise with standard deviation $ \lambda = 0.001$ and clip the gradients to have individual value magnitude of less than 0.01 similar to \citep{du2019implicit}.
The step size and number of steps for Langevin dynamics are different for different datasets as shown in Table~\ref{param}.

\begin{figure*}[t] 
\centering 
\centerline{\includegraphics[width=0.99\textwidth]{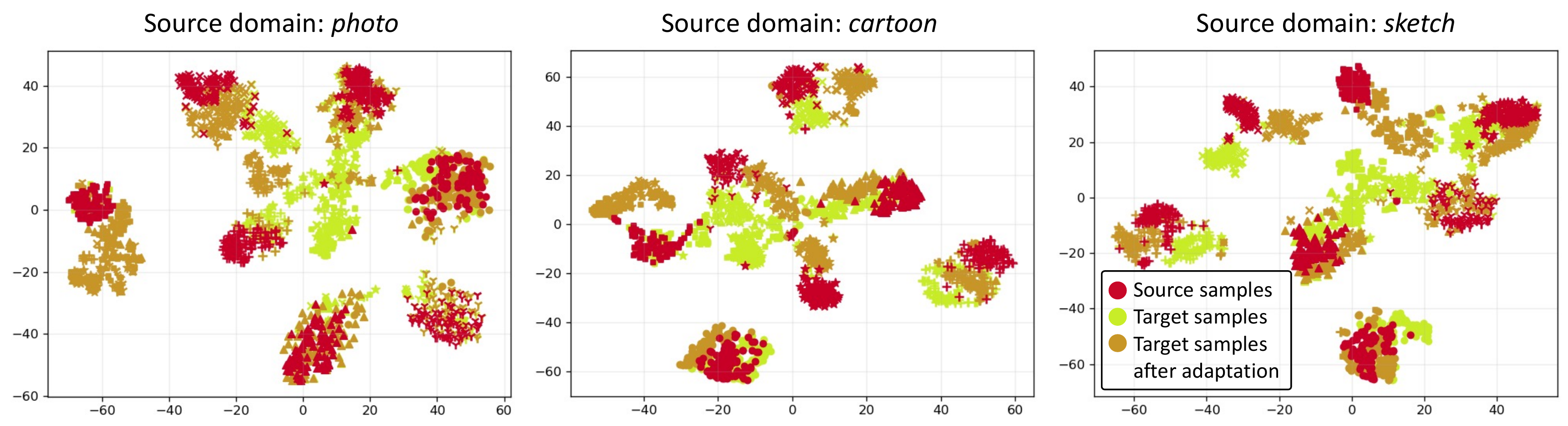}} 
\caption{\textbf{Benefit of energy-based test sample adaptation.} 
Different shapes denote different classes. From left to right: adaptation to the source domains \textit{photo}, \textit{cartoon}, and \textit{sketch} of samples from the target domain \textit{art-painting}. After adaptation, the target samples (\tikzsymbol[circle, mygrad, fill=mygrad]) are more close to the source data (\tikzsymbol[circle, myred, fill=myred]) than before (\tikzsymbol[circle, mygreen, fill=mygreen]), demonstrating the effectiveness of our method. Best viewed in color. 
} 
\label{visualization}
\vspace{-4mm}
\end{figure*} 

\begin{figure*}[t] 
\centering 
\centerline{\includegraphics[width=0.99\textwidth]{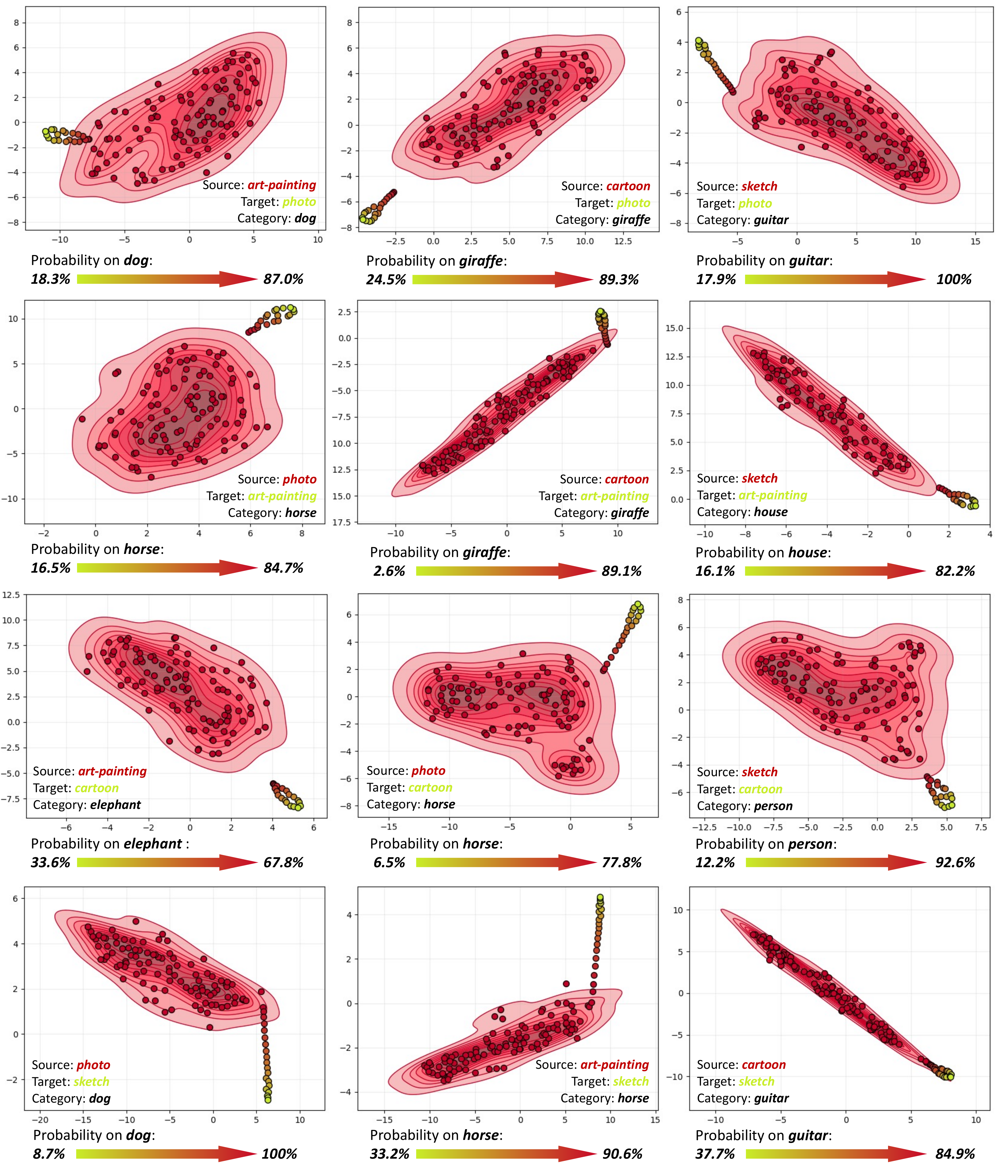}}
\caption{\textbf{More visualizations of the iterative adaptation on PACS.} We visualize the adaptation of samples from different target domains to source domains on PACS.
Each subfigure shows the adaptation of one target sample to one source domain. 
The adaptation procedure of the target samples is represented by the gradient color from green to red. 
In each figure, the target sample has the same label as the source data, but is mispredicted due to domain shifts.
During adaptation, the target samples gradually approach the source distributions and eventually get correct predictions.} 
\label{visualization2}
\vspace{-3mm}
\end{figure*} 

\section{Visualizations}
\label{app:vis}

\textbf{More visualizations of the adaptation procedure.}
To further show the effectiveness of the iterative adaptation of target samples, we provide more visualizations on PACS.
Figure~\ref{visualization} visualizes the source domain features and the target domain features both before and after the adaptation to each individual source domain.
Figure~\ref{visualization2} visualizes more iterative adaptation procedure of the target samples. Subfigures in different rows show the adaptation of samples from different target domains to source domains.
Similar with the visualizations in the main paper, the target samples gradually approach the source data distributions during the iterative adaptation.
Therefore, the predictions of the target samples on the source domain classifier become more accurate after adaptation.

\begin{figure*}[t] 
\centering 
\centerline{\includegraphics[width=0.99\textwidth]{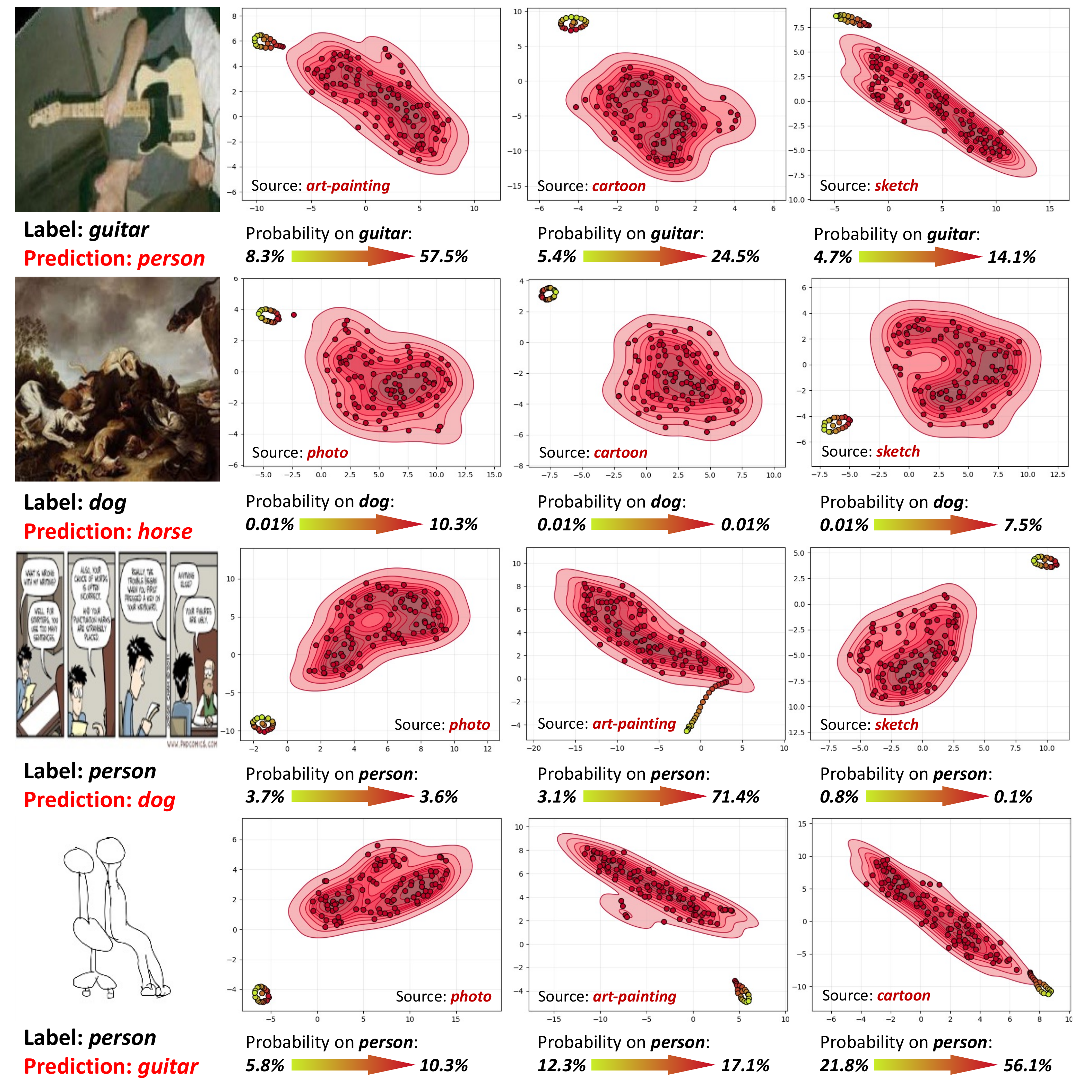}} 
\caption{\textbf{Failure case visualizations of our method on PACS.} 
The visualization settings are the same as Figure~\ref{visualization2}. Our method makes wrong predictions on samples with complex background or multiple objects. However, our method still achieves good adaptation of these target samples to some source domains, which shows its effectiveness.} 
\label{failure}
\vspace{-5mm}
\end{figure*}

\textbf{Failure cases.} 
We also provide some failure cases on PACS in Figure~\ref{failure} to gain more insights in our method.
Our method is confused with samples that have objects of different categories (first row) and multiple objects or complex background (last three rows).
A possible reason is that there is  noisy information contained in the latent variable of these samples, leading to adaptation without a clear direction, which behaves as wrong adaptation directions, e.g., visualization in row 1 column 4, or unstable updates with fluctuations in small regions, e.g., visualizations in row 2 column 3 and row 3 column 4.
Obtaining the latent variable with more accurate and clear categorical information can be one solution for these failure cases.
We can also solve the problem by achieving more stable adaptations with optimization techniques like Nesterov momentum \citep{nesterov1983method}.
Moreover, although failing in these cases, the adaptation of the target sample to some source domains still improves the performance, e.g., the adaptation of the \textit{photo} sample (row 1 column 2) and \textit{cartoon} sample (row 3 column 3) to the \textit{art-painting} domain and the adaptation of the \textit{sketch} sample (row 4 column 4) to the \textit{cartoon} domain, which further demonstrate the effectiveness of our iterative sample adaptation through the energy-based model.
The results motivate another solution for these failure cases, which is to learn to select the best source domain, or top-n source domains for adaptation and prediction of each target sample.
We leave these explorations for future work.

\section{More experimental results}
\label{app:results}

\begin{figure*}[t] 
\centering 
\centerline{\includegraphics[width=0.8\textwidth]{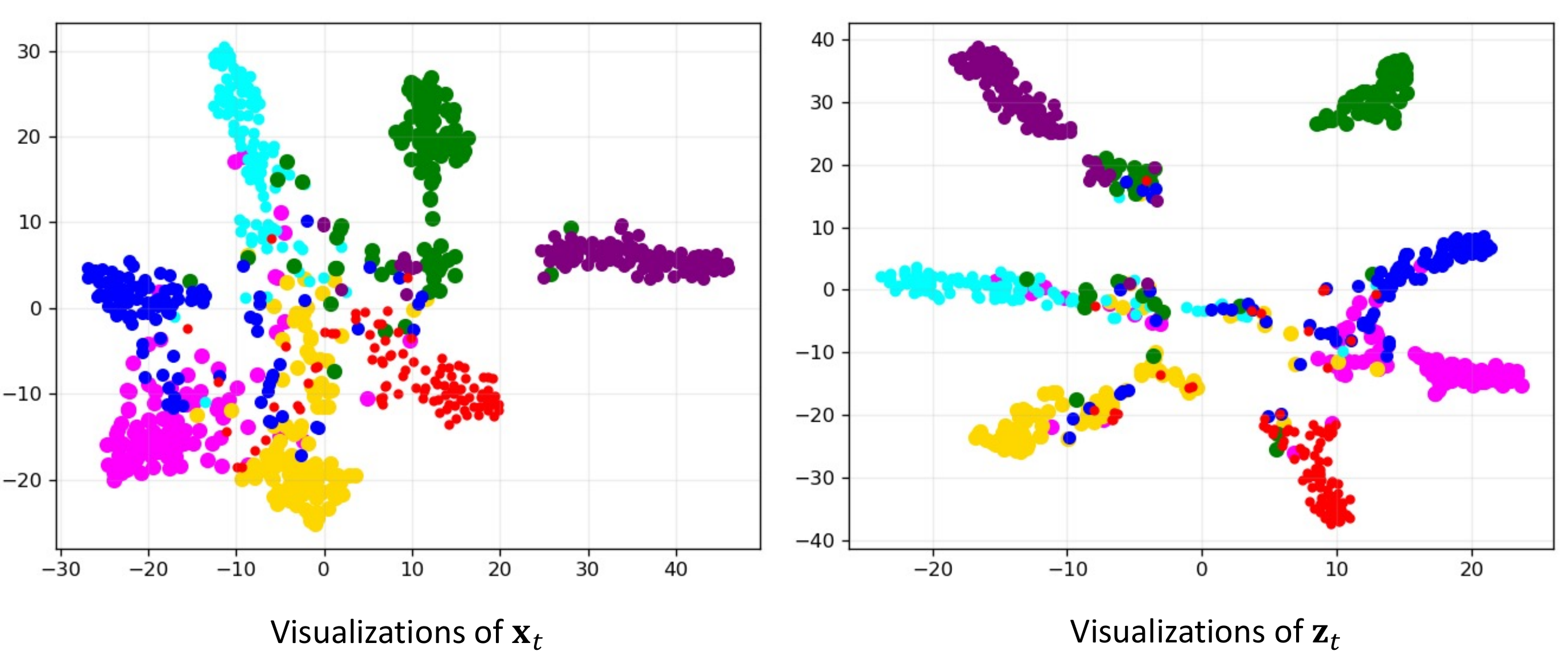}}
\caption{\textbf{Visualizations of the target features $\x_t$ and categorical latent variables $\z_t$.} We use \textit{art-painting}
on PACS as the target domain. Different colors denote different categories. $\z_t$ is obtained by $p_{\phi}(\z_t|\x_t)$. Most data points of $\z_t$ are clustered according to their labels, demonstrating that $\z_t$ can capture the categorical information.  } 
\label{visualize_xz}
\vspace{-3mm}
\end{figure*} 

\textbf{Analyses and discussions of the categorical latent variables.}
{
In the proposed method, the categorical latent representation for the test sample will have high fidelity to the correct class. 
This is guaranteed by the training procedure of our method. 
As shown in the training objective function (eq.~\ref{eq10explicitall}), we minimize the KL divergence to encourage the prior $p_{\phi}(\z|\x)$ to be close to the variational posterior $q_{\phi}(\z|\mathbf{d}_\x)$. $\mathbf{d}_\x$ is essentially the class prototype containing the categorical information. By doing so, we train the inference model $p_{\phi}(\z|\x)$ to learn to extract categorical information from a single sample.
Moreover, we also supervise the sample adaptation procedures by the predicted log-likelihood of the adapted samples (the last term in eq. (\ref{eq10explicitall})). The supervision is inherent in the objective function of our discriminative energy-based model as in the derivation of eq. (\ref{eq7:lall}) and eq. (\ref{eq10explicitall}). Due to this supervision, the model is trained to learn to adapt out-of-distribution samples to the source distribution while being able to maintain the correct categorical information conditioned on $\z$
Although trained only on source domains, the ability can be generalized to the target domain since it is trained by mimicking different domain shifts during training.}
To further show that $\z$ captures the categorical information in $\x$, we visualized the features $\x_t$ and latent variables $\z_t$ of the target samples in Figure~\ref{visualize_xz}, which shows that $\z_t$ actually captures the categorical information. 
Moreover, $\mathbf{z}_t$ is more discriminative than $\mathbf{x}_t$ as shown in the figure. 
Although $\z_t$ is approximated by only $\x_t$ during inference. 

{Moreover, the categorical latent variable benefits the correctness of the model in the case that the target samples are adapted to previously unexplored regions with very large numbers of steps.
Our optimization objective is to minimize the energy to adapt the sample, therefore it is possible that the energy of the target samples is lower than the source data after very large numbers of steps. 
In this case, the adapted samples could arrive in previously unexplored regions due to the limit of source data. 
This can further be demonstrated in Figure \ref{accene}, where the performance of the adapted samples drops after large numbers of steps, reaching a low energy value. Additionally, in the unexplored regions, the classifier could not be well trained, which might also be a reason for causing the performance drop. This is also one reason that we set the number of steps as a small value, e.g., 20 and 50.
The categorical latent variable benefits the correctness of the model in such cases as also can be found in Figure \ref{accene}. 
The oracle model shows almost no performance degradation even with small energy values after adaptation. 
The model with the latent variable $p(\z|\x)$ is also more robust to the step numbers and energy values than the model without $\z$. 
These results show the role of the latent variable in preserving the categorical information during adaptation and somewhat correcting prediction after adaptation.}

With the categorical latent variable $\z$, it is natural to make the final prediction directly by $p_{\phi}(\y|\z)$ without the sample adaptation procedure.
However, here we would like to clarify that it is sub-optimal.
The latent variable is dedicated to preserving the categorical information in $\mathbf{x}$ during adaptation. It still contains the domain information of the target samples. Therefore, it is not optimal to directly make predictions on the latent variable $\mathbf{z}$ due to the domain shifts between the $\mathbf{z}$ and the source-trained classifiers.
By contrast, the proposed method moves the target features close to the source distributions to address domain shifts while preserving the categorical information during adaptation.
To show how it works by direct prediction on $\mathbf{z}$, we provide the experimental results of making predictions only from $\mathbf{z}$ in Table~\ref{predonz}. As expected it is worse than predictions on the adapted target features $\mathbf{x}$, demonstrating the analysis we provided above.

\begin{table}[t]
\vspace{2mm}
\centering
\caption{{\textbf{
Analyses on the categorical latent variable.} As expected, prediction on the adapted target samples $\x$ performs better than prediction directly on the categorical latent variable $\z$.
}}
\label{predonz}
\begin{subtable}[t]{0.9\linewidth}
\centering
\caption{Overall comparisons on PACS and Office-Home.}
\centering
\label{zall}
	\resizebox{0.99\columnwidth}{!}{%
		\setlength\tabcolsep{4pt} 
\begin{tabular}{lllll}
\toprule
~ & \multicolumn{2}{c}{\textbf{PACS}} & \multicolumn{2}{c}{\textbf{Office-Home}} \\
\cmidrule(lr){2-3} \cmidrule(lr){4-5} 
~ & ResNet-18 & ResNet-50 & ResNet-18 & ResNet-50 \\ \midrule
predict directly on $\z$ & 82.46  \scriptsize{$\pm$0.34} & 85.95 \scriptsize{$\pm$0.33} & 64.49 \scriptsize{$\pm$0.25} & 70.60 \scriptsize{$\pm$0.53} \\
predict on adapted target samples $\x$  
& \textbf{84.92} \scriptsize{$\pm$0.59} & \textbf{87.70} \scriptsize{$\pm$0.28} & \textbf{66.31} \scriptsize{$\pm$0.21} & \textbf{72.07} \scriptsize{$\pm$0.38}\\ 
\bottomrule
\end{tabular}
}
\end{subtable}

\begin{subtable}[t]{0.99\linewidth}
\vspace{2mm}
\centering
\caption{{Detailed comparisons on PACS.}}
\centering
\label{zpacs}
	\resizebox{0.99\columnwidth}{!}{%
		\setlength\tabcolsep{4pt} 
\begin{tabular}{llllll}
\toprule
~  & \textbf{Photo}  & \textbf{Art-painting}         & \textbf{Cartoon}        & \textbf{Sketch}         & \textit{Mean}       \\ \midrule
\rowcolor{mColor2}
\multicolumn{6}{l}{\textbf{Predict directly on $\z$}}\\
No adaptation & 94,22 \scriptsize{$\pm$0.25} & 79.52 \scriptsize{$\pm$0.21} & 80.46 \scriptsize{$\pm$0.43} & 75.63 \scriptsize{$\pm$0.68} & 82.46 \scriptsize{$\pm$0.34} \\
\rowcolor{mColor2}
\multicolumn{6}{l}{\textbf{Predict on $\z$ with model adaptation (Tent)}}\\				
Adaptation with 1 sample per step & 80.49 \scriptsize{$\pm$0.27} & 44.14 \scriptsize{$\pm$0.38} & 51.49 \scriptsize{$\pm$0.44} & 30.28 \scriptsize{$\pm$0.66} & 51.60 \scriptsize{$\pm$0.37} \\
Adaptation with 16 samples per step & 93.65 \scriptsize{$\pm$0.33} & 80.20 \scriptsize{$\pm$0.24} & 76.90 \scriptsize{$\pm$0.52} & 68.49 \scriptsize{$\pm$0.72} & 79.81 \scriptsize{$\pm$0.31} \\
Adaptation with 64 samples per step & 96.04 \scriptsize{$\pm$0.33} & 81.91 \scriptsize{$\pm$0.37} & 80.81 \scriptsize{$\pm$0.64} & 76.33 \scriptsize{$\pm$0.65} & 83.77 \scriptsize{$\pm$0.41} \\
Adaptation with 128 samples per step & \textbf{97.25} \scriptsize{$\pm$0.24} & \textbf{84.91} \scriptsize{$\pm$0.31} & \textbf{81.12} \scriptsize{$\pm$0.47} & 76.80 \scriptsize{$\pm$0.83} & \textbf{85.02} \scriptsize{$\pm$0.49} \\
\rowcolor{mColor2}
\multicolumn{6}{l}{\textbf{Predict on adapted target samples $\x$ with our method}}\\
Adaptation with 1 sample $\x$ & 96.05 \scriptsize{$\pm$0.37} & 82.28 \scriptsize{$\pm$0.31} & \textbf{81.55} \scriptsize{$\pm$0.65} & \textbf{79.81} \scriptsize{$\pm$0.41} & \textbf{84.92} \scriptsize{$\pm$0.59} \\
\bottomrule
\end{tabular}
}
\end{subtable}
\end{table}

To show the advantages of our method, we also combine the prediction of the latent variable $\z$ with model adaptation methods.
We use the online adaptation proposed by \cite{wang2021tent}, where all target samples are utilized to adapt the source-trained models in an online manner. The model keeps updating step by step. In each step, the model is adapted to one batch of target samples.
As shown in Table~\ref{zpacs}, with large numbers of target samples per step, e.g., 128, the adaptation with Tent is competitive. However, when the number of samples for online adaptation is small, e.g., 1 and 16, the performance of the adapted model even drops, especially for single sample adaptation. By contrast, our method adapts each target sample to the source distribution. All target samples are adapted and predicted equally and individually. The overall performance of our method is comparable to Tent with 128 samples per adaptation step.

\begin{table}[t]
\caption{\textbf{Training cost of the proposed method.}
Compared with ERM, our method has about 20\% more parameters, most of which come from the energy functions of source domains. Similar to test time, the time cost of training increases along with the number of steps of the energy-based model. 
}
\centering
\label{traincost}
	\resizebox{0.7\columnwidth}{!}{%
		\setlength\tabcolsep{4pt} 
\begin{tabular}{lccc}
\toprule
 & Parameters & Adaptation steps & 10000 iterations training time  \\ \midrule
ERM & 11.18M  & - & ~6.2 h \\ \midrule
\multirow{5}*{\textbf{\textit{This paper}}} & \multirow{5}*{13.73M}  & ~20 & ~7.9 h \\
~ & ~ & ~40 & ~9.4 h \\ 
~ & ~ & ~60 & 10.6 h \\ 
~ & ~ & ~80 & 12.1 h \\ 
~ & ~ & 100 & 14.0 h \\
\bottomrule
\end{tabular}
}
\vspace{-2mm}
\end{table}

\textbf{Time cost with different adaptation steps.}
As the number of steps increases, both the training and test time cost consistently increases for all target domains. 
Without adaptation, the test time cost for one test batch is about 0.05 second.
The 20-step adaptation will take about 0.1 extra second. This number will increase to 0.25 second with 50 steps.
The test time increases by more than 0.5 seconds for 100 iterations, which is ten times that without adaptation and might limit the application of the proposal. 
The training time cost is more than two times that for ERM for 100 iterations as shown in Table~\ref{traincost}.
Since the extra time cost is mainly caused by the iterative adaptation, potential solutions can be speeding up the Langevin dynamics with some optimization techniques like Nesterov momentum \citep{nesterov1983method}, or exploring some one-step methods for the target sample adaptation. 
In other experiments on PACS in the paper we use 20 steps for all target domains considering both the overall performance and the time cost.



\begin{table}[t]
\vspace{-1mm}
\caption{\textbf{Comparisons on PACS.} Our method achieves best mean accuracy with a ResNet-18 backbone and is competitive with  ResNet-50. 
} 
\centering
\label{pacs}
	\resizebox{0.99\columnwidth}{!}{%
		\setlength\tabcolsep{4pt} 
\begin{tabular}{lllllll}
\toprule
Backbone & Method  & \textbf{Photo}    & \textbf{Art-painting}         & \textbf{Cartoon }       & \textbf{Sketch }        & \textit{Mean}       \\ \midrule
\multirow{7}*{ResNet-18}   & \cite{dou2019domain}   &  94.99   & 80.29                 & 77.17          & 71.69          & 81.04         \\
~  & \cite{iwasawa2021test} & -       & -  & -  & -  & 81.40  \\
~   & \cite{zhao2020domain}   & \textbf{96.65}          & 80.70          & 76.40          & 71.77          & 81.46         \\
~   & \cite{wang2021tent} &  95.49 & 81.55& 77.67  & 77.64 & 83.09       \\
~   & \cite{zhou2021mix}   & 96.10 & \textbf{84.10}          & 78.80          & 75.90          & 83.70 \\
~   & \cite{xiao2022learning} & 95.87       & 82.02  & 79.73  & 78.96  & 84.15  \\
~   & \textit{\textbf{This paper}}  & 96.05 \scriptsize{$\pm$0.37} & 82.28 \scriptsize{$\pm$0.31} & \textbf{81.55} \scriptsize{$\pm$0.65} & \textbf{79.81} \scriptsize{$\pm$0.41} & \textbf{84.92} \scriptsize{$\pm$0.59}\\
\midrule
\multirow{9}*{ResNet-50} & \cite{dou2019domain} & 95.01 & 82.89 & 80.49 & 72.29 & 82.67 \\
~ & \cite{dubey2021adaptive}  & -  & - & - & - & 84.50 \\
~   & \cite{iwasawa2021test} & -       & -  & -  & -  & 85.10  \\
~   & \cite{zhao2020domain} & \textbf{98.25} & 87.51 & 79.31 & 76.30 & 85.34 \\
~ & \cite{gulrajani2020search}  & 97.20 & 84.70 & 80.80 & 79.30 & 85.50 \\
~ & \cite{wang2021tent} & 97.96  & 86.30  & 82.53  & 78.11  & 86.23 \\
~   & \cite{seo2020learning} & 95.99 & 87.04 & 80.62 & \textbf{82.90} & 86.64 \\
~   & \cite{xiao2022learning}  & 97.88 & \textbf{88.09}  & 83.83  & 80.21    & \textbf{87.51} \\ 
~   & \textit{\textbf{This paper}}  & 97.67 \scriptsize{$\pm$0.14}         & \textbf{88.00} \scriptsize{$\pm$0.29}  & \textbf{84.75} \scriptsize{$\pm$0.39} & 80.40 \scriptsize{$\pm$0.38}         & \textbf{87.70} \scriptsize{$\pm$0.28} \\ 
\bottomrule
\end{tabular}
}
\end{table}

\begin{table}[t]
\begin{center}
\caption{\textbf{Comparisons on Office-Home.} Our method achieves the best mean accuracy using both a ResNet-18 and ResNet-50 backbone. 
} 
\label{office}
	\resizebox{0.99\columnwidth}{!}{%
		\setlength\tabcolsep{4pt} 
\begin{tabular}{lllllll}
\toprule
Backbone & Method  & \textbf{Art}    & \textbf{Clipart}         & \textbf{Product}       & \textbf{Real World}         & \textit{Mean}       \\ \midrule
\multirow{4}*{ResNet-18} & \cite{iwasawa2021test} & 47.00    & 46.80 & 68.00  & 66.10  & 57.00 \\
~ & \cite{wang2021tent} & 56.45 & 52.06 & 73.19  & 74.82  & 64.13
\\
~   & \cite{xiao2022learning}]  & 59.39  & \textbf{53.94}  & \textbf{74.68}  & 76.07  & 66.02  \\ 
~   & \textit{\textbf{This paper}}  & \textbf{60.08} \scriptsize{$\pm$0.33} & \textbf{53.93} \scriptsize{$\pm$0.34} & \textbf{74.50} \scriptsize{$\pm$0.39} & \textbf{76.74} \scriptsize{$\pm$0.24} & \textbf{66.31} \scriptsize{$\pm$0.21}\\
\midrule
\multirow{5}*{ResNet-50} & \cite{gulrajani2020search}  & 61.30 & 52.40 & 75.80 & 76.60 & 66.50 \\
~ & \cite{wang2021tent} & 62.12 & 56.65  & 75.61 & 77.58  & 67.99   \\
~ & \cite{dubey2021adaptive}  & -  & - & - & - & 68.90 \\
~   & \cite{xiao2022learning} & 67.21	 & 57.97 & 78.61 & 80.47 & 71.07 \\
~   & \textit{\textbf{This paper}}  & \textbf{69.33} \scriptsize{$\pm$0.14} & \textbf{58.37} \scriptsize{$\pm$0.30} & \textbf{79.29} \scriptsize{$\pm$0.32} & \textbf{81.26} \scriptsize{$\pm$0.26} & \textbf{72.07} \scriptsize{$\pm$0.38}\\
\bottomrule
\end{tabular}
}
\end{center}
\end{table}

\begin{table}[t]
\begin{minipage}[t]{\linewidth}
\vspace{-1mm}
\centering
\caption{\textbf{Generalization beyond image data.} Rumour detection on the PHEME microblog dataset. Our method achieves the best overall performance and is competitive in each domain. 
}
\centering
\label{text}
	\resizebox{0.99\columnwidth}{!}{%
		\setlength\tabcolsep{4pt} 
\begin{tabular}{lllllll}
\toprule
 & \textbf{Charlie Hebdo} & \textbf{Ferguson}    & \textbf{German Wings }        & \textbf{Ottawa Shooting  }      & \textbf{Sydney Siege}        & \textit{Mean}       \\ \midrule
ERM Baseline & 79.4 \scriptsize{$\pm$0.25} & 77.1 \scriptsize{$\pm$0.36} & \textbf{75.7} \scriptsize{$\pm$0.12} & 68.2 \scriptsize{$\pm$0.48} & 75.0 \scriptsize{$\pm$0.28} & 75.1 \scriptsize{$\pm$0.29} \\
\cite{wang2021tent} & 80.1 \scriptsize{$\pm$0.18} & 76.9 \scriptsize{$\pm$0.56} & 74.7 \scriptsize{$\pm$0.52} & \textbf{72.0} \scriptsize{$\pm$0.48} & 75.4 \scriptsize{$\pm$0.34} & 75.8 \scriptsize{$\pm$0.23} \\
\cite{xiao2022learning} & 81.0 \scriptsize{$\pm$0.52} & 77.2 \scriptsize{$\pm$0.25} & \textbf{75.7} \scriptsize{$\pm$0.31} & 70.0 \scriptsize{$\pm$0.46} & \textbf{76.5} \scriptsize{$\pm$0.22} & 76.1 \scriptsize{$\pm$0.21} \\
\textit{\textbf{This paper}} & \textbf{81.8} \scriptsize{$\pm$0.43} & \textbf{77.8} \scriptsize{$\pm$0.37}& 75.3 \scriptsize{$\pm$0.14} & \textbf{71.9} \scriptsize{$\pm$0.28} & 75.6 \scriptsize{$\pm$0.15} & \textbf{76.5} \scriptsize{$\pm$0.18}\\
\bottomrule
\end{tabular}
}
\end{minipage}
%
%
\begin{minipage}[t]{\linewidth}
\vspace{2mm}
\centering
\caption{
\textbf{
Experiments on single-source domain generalization.} The model is trained on original data and evaluated on 15 different types of corruption. Our method is competitive with \cite{sun2020test}, \cite{rusak2020simple} and \cite{hendrycks2020augmix}, and is outperformed by \cite{wang2021tent}. Mimicking good domain shifts during training is important for our method.}
\vspace{-1mm}
\centering
\label{ssdg}
	\resizebox{0.7\columnwidth}{!}{%
		\setlength\tabcolsep{4pt} 
\begin{tabular}{lcc}
\toprule
Method  & \textbf{CIFAR-10-C} & \textbf{ImageNet-C}  \\ \midrule
\cite{yun2019cutmix}
& {31.1} & -\\
\cite{guo2019mixup}
& {25.8} & -\\
\cite{hendrycks2020augmix}
& {17.4} & 51.7\\
\cite{rusak2020simple}
& - & 50.2\\
\cite{sun2020test}
& 17.5 & -\\
\cite{wang2021tent}
& \textbf{14.3} & \textbf{44.0}\\ 
\textit{\textbf{This paper}} (noisy data as negative samples)
& 21.5 & 55.8\\ 
\textit{\textbf{This paper}} (corrupted data as negative samples)
& 17.0 & 51.1\\ 
\bottomrule
\end{tabular}
}
\vspace{-6mm}
\end{minipage}
%
%
\end{table}

\textbf{Detailed comparisons.}
We provide the detailed performance of each target domain on PACS (Table~\ref{pacs}), Office-Home (Table~\ref{office}), and PHEME (Table~\ref{text}).
On PACS, our method achieves competitive results on each target domain and the best overall performance with both ResNet-18 and ResNet-50 as the backbone.
Moreover, our method performs better than most of the recent model adaptation methods \citep{wang2021tent,dubey2021adaptive,iwasawa2021test,xiao2022learning}.
The conclusion on Office-Home and PHEME is similar to that on PACS. We achieve competitive and even better performance on each target domain.


\textbf{Results on single source domain generalization.}
To show the ability of our method of handling corruption distribution shifts and single source domain generalization, we conduct some experiments on CIFAR-10-C and ImageNet-C. We train the model on original data and evaluate it on the data with 15 types of corruption. 

Since our method needs to mimic distribution shifts to train the discriminative energy-based model during training, for the single source domain setting, we generate the negative samples by adding random noise to the image and features of the clean data.  We also use the other 4 corruption types (not contained in the evaluation corruption types) as the negative samples during training, which we regard as “corrupted data as negative samples”. Note that these corrupted data are only used as negative samples to train the energy-based model.
As shown in Table~\ref{ssdg}, by mimicking better domain shifts during training, our method achieves competitive results with \cite{sun2020test}. 
{
We also compare our method with some data-augmentation-based methods (e.g., Mixup \citep{guo2019mixup}, CutMix \citep{yun2019cutmix} and AugMix \citep{hendrycks2020augmix}), our sample adaptation is also competitive with these methods.}
The proposed method performs worse with a single source domain, although we generate extra negative samples to mimic the domain shifts. The reason can be that the randomly generated domain shifts do not well simulate the domain shift at test time. 

{\textbf{Analyses for ensemble prediction.}}
{We conduct several experiments on PACS to analyze the ensemble inference in our method.
We first provide the results of each source-domain-specific classifier before and after sample adaptation. As shown in Table \ref{persource}, Although the performance before and after adaptation to different source domains are different due to domain shifts, the proposed sample adaptation to most of the source domains performs better. 
Moreover, the ensemble inference further improves the overall performance of both without and with sample adaptation, where the results with sample adaptation are still better, as expected.}

We also try different aggregation methods to make the final predictions. The results are provided in Table~\ref{simpred}. 
The best results in Table \ref{persource} are comparable, but it is difficult to find the best source domain for adaptation before obtaining the results. 
We tried to find the closest source domain of each target sample by the cosine similarity of feature representations and the predicted confidence. We also tried to aggregate the predictions by weighted average according to the Cosine similarity. 
With cosine similarity, the weighted averaged results are comparable to the common ensemble method we used in the paper, but the results of adaptation to the closest source domain are not so good. 
The reason can be that the cosine measure is not able to estimate the domain relationships, showing that it is difficult to reliably estimate the relationships between source and single target samples.
The results obtained by using the most confident adaptation are also not as good as the ensemble method, although comparable. The reason can be that ensemble methods introduce uncertainty into the predictions, which is more robust.

\begin{table}[t]
\vspace{2mm}
\centering
\caption{{\textbf{
Sample adaptation to each source domain on PACS.} The experiments are conducted on ResNet-18. Due to the domain shifts between the target domain and different source domains, the performance before and after adaptation are different. The results with sample adaptation to most of the source domains are better than those without adaptation. The ensemble inference further improves the overall performance, where the results with sample adaptation are still better than that without adaptation.
}}
\label{persource}
\begin{subtable}[t]{0.495\linewidth}
\vspace{2mm}
\centering
\caption{{Photo}}
\centering
\label{pspho}
	\resizebox{0.99\columnwidth}{!}{%
		\setlength\tabcolsep{4pt} 
\begin{tabular}{llllll}
\toprule
~  & Art-painting         & Cartoon        & Sketch         & \textit{Ensemble}       \\ \midrule
w/o adaptation
&95.79 \scriptsize{$\pm$0.23} & 95.03 \scriptsize{$\pm$0.27} & 95.05 \scriptsize{$\pm$0.42} & 95.12 \scriptsize{$\pm$0.41} \\
w/ adaptation
& 95.81 \scriptsize{$\pm$0.27} & 94.69 \scriptsize{$\pm$0.21}  & 95.99 \scriptsize{$\pm$0.45} & 96.05 \scriptsize{$\pm$0.37} \\ 
\bottomrule
\end{tabular}
}
\end{subtable}
\begin{subtable}[t]{0.495\linewidth}
\vspace{2mm}
\centering
\caption{{Art-painting}}
\centering
\label{psart}
	\resizebox{0.99\columnwidth}{!}{%
		\setlength\tabcolsep{4pt} 
\begin{tabular}{llllll}
\toprule
~  & Photo         & Cartoon        & Sketch         & \textit{Ensemble}       \\ \midrule
w/o adaptation
&78.52 \scriptsize{$\pm$0.43} & 79.68 \scriptsize{$\pm$0.37} & 79.83 \scriptsize{$\pm$0.52} & 79.79 \scriptsize{$\pm$0.64} \\
w/ adaptation
& 81.49 \scriptsize{$\pm$0.33} & 82.19 \scriptsize{$\pm$0.35}  & 80.81 \scriptsize{$\pm$0.43} & 82.28 \scriptsize{$\pm$0.31} \\ 
\bottomrule
\end{tabular}
}
\end{subtable}

\begin{subtable}[t]{0.495\linewidth}
\vspace{2mm}
\centering
\caption{{Cartoon}}
\centering
\label{pscart}
	\resizebox{0.99\columnwidth}{!}{%
		\setlength\tabcolsep{4pt} 
\begin{tabular}{llllll}
\toprule
~  &  Photo   & Art-painting            & Sketch         & \textit{Ensemble}       \\ \midrule
w/o adaptation
&79.05 \scriptsize{$\pm$0.33} & 78.93 \scriptsize{$\pm$0.41} & 78.80 \scriptsize{$\pm$0.55} & 79.15 \scriptsize{$\pm$0.37} \\
w/ adaptation
& 81.09 \scriptsize{$\pm$0.38} & 80.44 \scriptsize{$\pm$0.31}  & 80.32 \scriptsize{$\pm$0.71} & 81.55 \scriptsize{$\pm$0.65} \\ 
\bottomrule
\end{tabular}
}
\end{subtable}
\begin{subtable}[t]{0.495\linewidth}
\vspace{2mm}
\centering
\caption{{Sketch}}
\centering
\label{pssket}
	\resizebox{0.99\columnwidth}{!}{%
		\setlength\tabcolsep{4pt} 
\begin{tabular}{llllll}
\toprule
~  & Photo         & Art-painting            & Cartoon         & \textit{Ensemble}       \\ \midrule
w/o adaptation
&78.32 \scriptsize{$\pm$0.56} & 76.98 \scriptsize{$\pm$0.73} & 76.16 \scriptsize{$\pm$0.82} & 79.28 \scriptsize{$\pm$0.82} \\
w/ adaptation
& 79.72 \scriptsize{$\pm$0.43} & 79.69 \scriptsize{$\pm$0.67}  & 79.77 \scriptsize{$\pm$0.45} & 79.81 \scriptsize{$\pm$0.41} \\ 
\bottomrule
\end{tabular}
}
\end{subtable}
\end{table}

\begin{table}[t]
\vspace{2mm}
\centering
\caption{{
\textbf{Analyses of different aggregation methods for the predictions.} The experiments are conducted on PACS using ResNet-18. The results with different aggregation methods are similar while ensemble inference performs slightly better.
}}
\centering
\label{simpred}
	\resizebox{0.99\columnwidth}{!}{%
		\setlength\tabcolsep{4pt} 
\begin{tabular}{llllll}
\toprule
Aggregation methods  & Photo    & Art-painting         & Cartoon        & Sketch         & \textit{Mean}       \\ \midrule
Adaptation to the closest source domain
&95.41 \scriptsize{$\pm$0.28} & 79.86 \scriptsize{$\pm$0.41} & 79.67 \scriptsize{$\pm$0.44} & 78,97 \scriptsize{$\pm$0.72} & 83.48 \scriptsize{$\pm$0.43} \\
Weighted average of adaptation to different source domains
&95.93 \scriptsize{$\pm$0.33} & 82.18 \scriptsize{$\pm$0.37} & 81.24 \scriptsize{$\pm$0.52} & 79.54 \scriptsize{$\pm$0.77} & 84.76 \scriptsize{$\pm$0.55} \\
Most confident prediction after adaptation
&95.77 \scriptsize{$\pm$0.25} & 81.93 \scriptsize{$\pm$0.31} & 80.67 \scriptsize{$\pm$0.65} & 79.25 \scriptsize{$\pm$0.62} & 84.41 \scriptsize{$\pm$0.32} \\
Ensemble (This paper)
& 96.05 \scriptsize{$\pm$0.37} & {82.28} \scriptsize{$\pm$0.31}  & {81.55} \scriptsize{$\pm$0.65} & {79.81} \scriptsize{$\pm$0.41} & {84.92} \scriptsize{$\pm$0.59} \\ 
\bottomrule
\end{tabular}
}
\vspace{-6mm}
\end{table}

\begin{figure*}[t!] 
\centering 
\centerline{\includegraphics[width=0.99\textwidth]{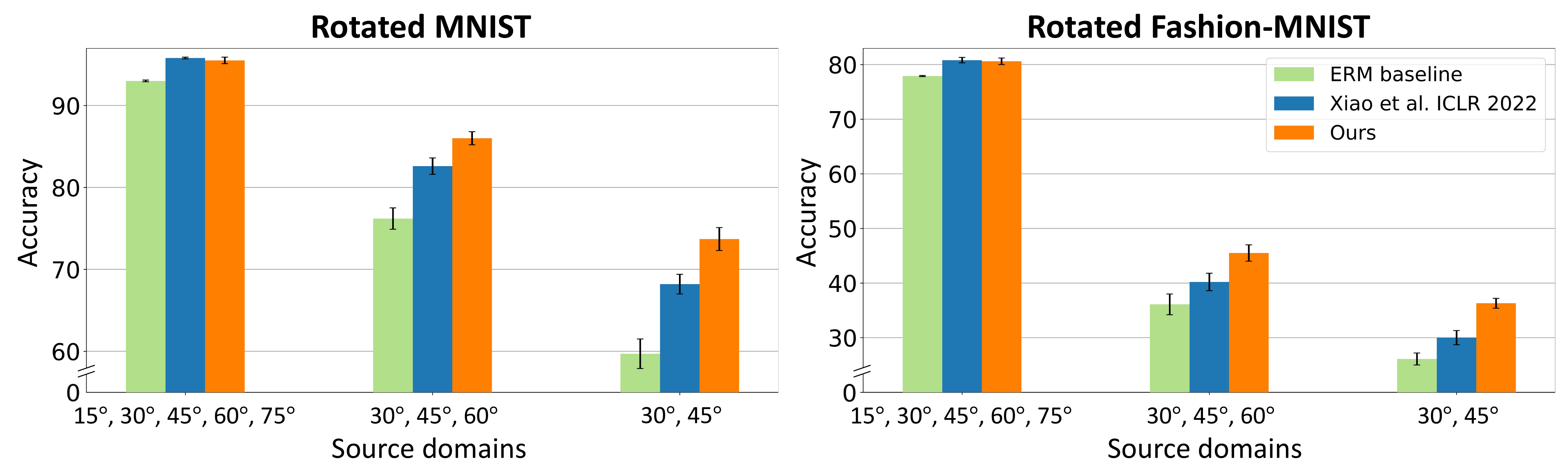}} 
\caption{\textbf{Benefit for larger domain gaps.} 
We train on different source set distributions and evaluate on target sets with rotation angles of $0^\circ$ and $90^\circ$. As the domain gap between source and target sets increases, our method performs better than alternatives. 
} 
\label{mnist}
\end{figure*}

\textbf{Benefit for larger domain gaps.}
To show the benefit of our proposal for domain generalization scenarios with large gaps, we conduct experiments on rotated MNIST and Fashion MNIST.
The results are shown in Figure~\ref{mnist}.
The models are trained on source domains with rotation angles from $15^\circ$ to $75^\circ$, $30^\circ$ to $60^\circ$, and $30^\circ$ to $45^\circ$, and always tested on target domains with angles of $0^\circ$ and $90^\circ$.
As the number of domains seen during training decreases the domain gap between source and target increases, and the performance gaps between our method and others becomes more pronounced.
Notably, when comparing our method with the recent test-time adaptation of \cite{xiao2022learning}, which adapts a model to each target sample, shows adapting target samples better handles larger domain gaps than adapting the model.

\end{document}